\documentclass[preprint,12pt]{elsarticle}

\usepackage{lineno}
\usepackage{hyperref}
\usepackage{amsmath}
\usepackage{graphicx}
\usepackage{tabularx} 
\usepackage{makecell}   
\usepackage{array}       
\usepackage{tabularx}
\usepackage{ragged2e}
\modulolinenumbers[5]

\journal{AI Open}

\begin{document}
	
	\begin{frontmatter}
		
		\title{Personalized Artificial General Intelligence (AGI) via Neuroscience-Inspired Continuous Learning Systems}
		
		\author[1]{Rajeev Gupta}
		\author[2]{Suhani Gupta}
		\author[3]{Ronak Parikh}
		\author[4]{Divya Gupta}
		\author[5]{Amir Javaheri}
		\author[6]{Jairaj Singh Shaktawat}
		
		\address[1]{Department of AI and Robotics, IGNITE Pathways, Dublin, California, USA}
		\address[2]{Department of Youth Wellness and Research, Neuro Health Alliance, Dublin, California, USA}
		\address[3]{Independent Researcher, New York, New York, USA}
		\address[4]{Medical Director of JFK Neuroscience Sleep Center, HMH-JFK Neuroscience Institute, Edison, New Jersey, USA}
		\address[5]{Independent Researcher, Orinda, California, USA}
		\address[6]{Independent Researcher, Toronto, CA}
				
		\begin{abstract}
			Artificial Intelligence has made remarkable advancements in recent years, primarily driven by increasingly large deep learning models. However, achieving true Artificial General Intelligence (AGI) demands fundamentally new architectures rather than merely scaling up existing models. Current approaches largely depend on expanding model parameters, which improves task-specific performance but falls short in enabling continuous, adaptable, and generalized learning. Achieving AGI capable of continuous learning and personalization on resource-constrained edge devices is an even bigger challenge.
			
			This paper reviews the state of continual learning and neuroscience-inspired AI, and proposes a novel architecture for Personalized AGI that integrates brain-like learning mechanisms for edge deployment. We review literature on continuous lifelong learning, catastrophic forgetting, and edge AI, and discuss key neuroscience principles of human learning, including Synaptic Pruning, Hebbian plasticity, Sparse Coding, and Dual Memory Systems, as inspirations for AI systems. Building on these insights, we outline an AI architecture that features complementary fast-and-slow learning modules, synaptic self-optimization, and memory-efficient model updates to support on-device lifelong adaptation.
			
			Conceptual diagrams of the proposed architecture and learning processes are provided. We address challenges such as catastrophic forgetting, memory efficiency, and system scalability, and present application scenarios for mobile AI assistants and embodied AI systems like humanoid robots. We conclude with key takeaways and future research directions toward truly continual, personalized AGI on the edge. While the architecture is theoretical, it synthesizes diverse findings and offers a roadmap for future implementation.
		\end{abstract}
		
		
	\end{frontmatter}

	\section*{Acknowledgements}
	We would like to thank Ronav Gupta, Founder and President of IGNITE Pathways, for his thoughtful feedback throughout the development of this paper. His insights on neuroscience-inspired learning and system design greatly enhanced the clarity and depth of the work.
	
	Additionally, we acknowledge the use of GPT-4 for assistance with refining language and structure throughout the drafting process, and MidJourney’s diffusion-based image generation model for support in generating illustrative conceptual figures. These tools supported clarity and visualization while preserving the integrity and originality of the ideas presented.
	
	\section{Introduction}
	Artificial General Intelligence (AGI) is often envisioned as an AI system with versatile, human-like cognitive abilities that can learn and adapt over a lifetime. One critical aspect of human intelligence is continuous learning, the ability to accumulate knowledge and skills over time without forgetting previously learned abilities. For AI to approach human-level adaptability, it must learn continually from new experiences, personalize to specific users or environments, and do so efficiently under real-world constraints. Achieving this on edge devices (such as humanoids, IoT gadgets, and autonomous robots) is particularly challenging due to limited computational resources and the need for on-device processing (for privacy, latency, or reliability reasons). Current AI systems, however, typically learn in static offline training phases and are then deployed as fixed models that do not evolve. When confronted with novel situations, they often require complete retraining or cloud-based updates, which is impractical for lifelong personalization~\cite{johnson2024edgeai}.
	
	Recent research highlights the gap between biological continuous learning and today’s AI. No single AI model currently exists that incorporates all the features of lifelong learning observed in biological brains~\cite{johnson2024edgeai}. Humans learn continuously by employing a variety of neural mechanisms, including adjusting synaptic connections on the fly, encoding sparse efficient representations, consolidating important memories, and pruning redundant ones.
	
	Neuroscience-inspired AI seeks to bridge this gap by translating principles of brain learning into machine learning algorithms. For instance, biological brains achieve remarkable lifelong learning with tiny energy budgets, the human brain uses approximately 20 W for trillions of synapses~\cite{johnson2024edgeai,laughlin2003communication}, suggesting there are highly efficient learning and memory management strategies at play that AI could emulate. Leveraging such techniques may enable AI systems that can learn continuously on the edge without catastrophic failures or resource exhaustion.
	
	This paper introduces a novel, neuroscience-inspired AI architecture that enables lifelong, personalized learning on edge devices. Our approach integrates complementary memory systems for both fast and slow learning with dynamic model compression techniques, such as pruning, distillation, and sparse activation, to overcome catastrophic forgetting and optimize on-device resource usage. 
	
	In the following sections, we first review the state-of-the-art in continual learning and neuroscience-inspired AI, highlighting both the challenges and promising strategies in this rapidly evolving field. Next, we delve into neuroscience principles that underlie human-like learning, Synaptic Pruning, Hebbian Plasticity, Sparse Coding, and Dual Memory Systems, explaining each and discussing how they can inform AI algorithms. We then propose an AI architecture for personalized AGI that integrates these principles. The architecture is designed to support continuous adaptation by combining complementary memory systems (for fast and slow learning) with model compression techniques (pruning, distillation, sparse activation) to remain efficient on edge devices.
	
	A methodology for evaluating the proposed system is outlined, including potential datasets and benchmarks to assess continual learning performance and adaptability. We include conceptual figures to illustrate the architecture and key mechanisms for clarity. In the discussion, we address practical challenges such as catastrophic forgetting, memory efficiency, and the scalability of the approach as the system learns more over time. We also describe application scenarios where such an on-device continual learning AGI would be valuable, from autonomous humanoids that learn in dynamic environments to robots and mobile assistants that personalize to user behaviors. Finally, we summarize the key findings and highlight future research directions toward realizing neuroscience-guided lifelong learning in AGI systems on the edge.
	
	\section{Literature Review}
	
	\subsection{Continuous Learning and Catastrophic Forgetting in AI}
	Traditional machine learning models are trained on fixed datasets and do not naturally handle sequential learning of multiple tasks. In contrast, a continual learning (or lifelong learning) paradigm involves learning a stream of tasks or data distributions over time. A well-known pitfall in this setting is catastrophic forgetting, when learning new tasks causes a model to lose performance on previously learned tasks. Early studies by Goodfellow et al.~\cite{goodfellow2015forgetting} provided empirical evidence of catastrophic forgetting in neural networks, showing that after sequential training, models tend to “forget” how to perform earlier tasks. They found that no standard optimization could completely eliminate forgetting, though techniques like dropout mitigated it to some extent. This highlighted the stability–plasticity dilemma: the need to learn new information (plasticity) while retaining old knowledge (stability).
	
	Over the past decade, many techniques have been proposed to address catastrophic forgetting. These can be grouped into: (1) regularization-based methods, (2) memory replay methods, and (3) dynamic architecture methods~\cite{johnson2024edgeai}.
	
	Regularization approaches add constraints to the training objective to protect important weights. For example, Elastic Weight Consolidation computes a Fisher-information based importance ~\cite{kirkpatrick2017ewc} for each parameter and penalizes changes to important parameters when learning new tasks. Similarly, Synaptic Intelligence (SI)~\cite{zenke2017si} accumulates an importance measure based on contribution to performance. These methods are inspired by the idea of synaptic consolidation in the brain, where important synapses become less plastic to preserve long-term memories. However, purely regularization-based methods can struggle when tasks are very different, as they inevitably must compromise plasticity for stability.
	
	Memory replay methods maintain a subset of past training data or a generative model of past experiences to rehearse previously learned tasks while learning new ones. For instance, experience replay stores a small buffer of past samples and interleaves them with new task data during training~\cite{robins1995replay}. This strategy is loosely motivated by the hippocampal replay observed in biological brains during sleep, believed to consolidate memories. Modern variants include iCaRL~\cite{rebuffi2017icarl}, which stores examples per class for incremental class learning, and generative replay~\cite{shin2017generative}, which trains a generative network to synthesize pseudo-data from past tasks for rehearsal. Brain-inspired versions of replay have also been explored: van de Ven et al.~\cite{vandeven2020brainreplay} introduced a neurologically motivated replay mechanism that improved continual learning by mimicking how the brain reactivates neural patterns during memory consolidation. Replay methods significantly alleviate forgetting, but storing data can be memory-intensive and raises privacy concerns (which is problematic for personalized on-device learning), whereas generative replay adds computational overhead.
	
	A recent and comprehensive empirical survey by Mai et al. (2021)~\cite{mai2021online}  provides a valuable comparison of state-of-the-art online continual learning (OCL) methods for image classification. It evaluates approaches such as MIR, GDumb, iCaRL, and ER under various scenarios, including class-incremental and domain-incremental settings. The study also emphasizes the importance of practical factors such as memory buffer size, task ordering, and the choice of classifiers (e.g., nearest class mean vs. softmax). By systematically analyzing OCL techniques, the paper underscores current challenges in balancing accuracy, memory constraints, and computational efficiency, issues that are central to edge-deployable, lifelong learning systems. The findings reinforce the need for flexible yet efficient architectures capable of robust performance in real-world, streaming data environments.
	
	Dynamic architecture methods tackle continuous learning by expanding or reconfiguring the model’s architecture to accommodate new knowledge. One example is Progressive Neural Networks~\cite{rusu2016progressive}, which allocate a new neural network column for each task and leverage lateral connections to previously learned features. This completely avoids interference by design (each task uses its own parameters), but the model grows linearly with the number of tasks, which is not feasible for edge deployment. Later work has aimed for more compact expansion: Dynamic Expandable Networks~\cite{yoon2018den} grow neural units only as needed based on loss improvement, and PackNet~\cite{mallya2018packnet} uses iterative pruning to free up network capacity for new tasks, “packing” multiple tasks into a single model by assigning disjoint parameter subsets. Such approaches connect to neuroscience via concepts of neurogenesis (adding neurons) and synaptic pruning (removing connections), which we will discuss in depth later. Recent research in lifelong learning accelerators emphasizes the value of reconfigurable architectures that can add neurons or synapses to learn new knowledge while preserving old knowledge in-place~\cite{johnson2024edgeai}. For example, adding “extra” neurons or layers to a network for new information can mitigate forgetting by isolating new learning to those additions. These dynamic expansion methods increase model size by allocating more parameters (e.g., an entire “column” per task in Progressive Neural Networks), and even more efficient approaches like Dynamic Expandable Networks still grow two to three fold over multiple tasks. For instance, a base network of 1.6 million parameters can balloon to over 16 million parameters (10$\times$) when learning 10 tasks with Progressive Neural Networks, an obvious non-starter for edge devices with limited memory. Although methods such as PackNet mitigate parameter growth by pruning and reusing weights, they introduce storage overheads for pruning masks. Consequently, any approach that relies on adding neurons or layers is generally not feasible in memory-constrained or edge scenarios.

	\subsection{Neuroscience-Inspired AI Approaches}
	Neuroscience has long been a source of inspiration for AI, from the early days of neural networks (inspired by brain neurons) to current research in lifelong learning. Neuroscience-inspired AI can refer both to algorithms that incorporate brain-like mechanisms and to hardware that emulates neural processes (neuromorphic computing). Here we focus on algorithmic inspirations relevant to continual learning and personalization.
	
	One key insight from neuroscience is that biological brains employ multiple interacting learning systems. For example, the Complementary Learning Systems (CLS) theory~\cite{mcclelland1995cls} posits that the brain’s memory is supported by a fast-learning but temporary store (the hippocampus) and a slow-learning long-term store (the neocortex). This has inspired AI architectures that separate fast adaptation from slow knowledge accumulation. Recent work by Wang et al.~\cite{wang2023neuroinspired} drew from a Drosophila (fruit fly) learning model to design a multi-module AI system that actively regulates forgetting. In flies, different neural modules handle parallel memory traces. Some memories are allowed to decay if they prove irrelevant, to maintain plasticity for new learning. Wang et al. mimic this by coordinating multiple learners: old memories are attenuated in one part of the network to improve plasticity for new information, while another part ensures important knowledge is retained. This approach outperformed traditional synaptic regularization methods (like EWC) in continual learning benchmarks, underscoring how bio-inspired strategies (in this case, purposeful forgetting in certain modules) can improve flexibility.
	
	Another area of neuroscience inspiration is in learning rules. The classical Hebbian learning rule (“cells that fire together wire together”) describes how synaptic connections strengthen when pre- and postsynaptic neurons are co-activated. While standard deep learning uses gradient backpropagation rather than local Hebbian updates, there is growing interest in incorporating Hebbian-style plasticity for rapid adaptation. For example, fast weights approaches~\cite{schmidhuber1992,ba2016fastweights} treat certain weights as dynamically updatable by Hebbian rules to store recent context. Munkhdalai et al.~\cite{munkhdalai2019metalearning} and others have explored meta-learning algorithms where some connections are learned to adapt quickly in a Hebbian manner to new data. The recent “Titans” architecture by Behrouz et al.~\cite{behrouz2025titans} provides a practical instantiation: it combines a standard attention-based Transformer (interpreted as a short-term memory for current context) with an external learned memory module that serves as long-term memory. In their view, attention mechanisms with limited context can play the role of working memory, while a differentiable memory captures long-range dependencies as a persistent store. Such architectures echo the idea of a dual memory system and use learning rules that allow fast incorporation of information into the memory module (potentially implementable with local Hebbian-like updates). Incorporating these ideas, AI systems can adjust certain parameters rapidly during inference or on the fly, enabling one-shot or few-shot learning of new patterns, analogous to how humans can quickly memorize a new fact.
	
	Neuromorphic computing is another facet of neuroscience-inspired AI, targeting the deployment side. Neuromorphic chips (such as Intel Loihi or IBM TrueNorth) use spiking neural network models and event-driven operation to mimic the brain’s hardware efficiency. These chips support on-chip learning with local plasticity rules (e.g., spike-timing dependent plasticity, a form of Hebbian learning) and operate with extremely low power, making them promising for edge devices. For instance, the startup BrainChip’s Akida processor implements spiking neurons with on-chip learning for vision tasks, directly inspired by neurobiology~\cite{johnson2024edgeai}. While spiking networks are not yet as accurate as deep learning on complex tasks, research prototypes have shown the feasibility of incremental learning on devices with low energy use~\cite{davies2018loihi}.
	
	Additionally, the concept of memory consolidation during sleep has inspired algorithms that simulate an “offline” phase (which could be when the edge device is idle or charging) to replay memories or reorganize knowledge, similar to how the brain consolidates learning during sleep cycles.
	
	\subsection{Edge AI Deployment and Model Compression}
	Deploying AGI-like capabilities on edge devices comes with constraints in model size, compute, and energy. Modern state-of-the-art AI models, especially in natural language or vision, are far too large for direct on-device use and are typically trained on powerful cloud servers. To enable on-device inference, researchers employ model compression and efficiency techniques.
	
	Model distillation is a common approach where a large model’s knowledge is transferred to a smaller model by training the small “student” model to mimic the outputs (or intermediate representations) of the large “teacher” model~\cite{hinton2015distilling}. Distillation has been used to compress models for mobile deployment while retaining high accuracy. For example, powerful cloud-trained models could be distilled into a personalized model on the device periodically, providing the device with a strong prior knowledge base that it can further personalize with local learning.
	
	Another technique is quantization, which reduces the numerical precision of model parameters (e.g., 8-bit integers instead of 32-bit floats) to shrink memory footprint and accelerate inference on hardware that supports low-precision arithmetic. Many edge devices now support 8-bit or mixed-precision operations, and research is progressing toward training and inference with even lower precision (down to 4-bit or binary networks) to save energy. However, as discussed in ~\cite{johnson2024edgeai}, referencing Kudithipudi et al., today’s deployed neural networks often run at reduced precision (8-bit), which is sufficient for fixed inference but not for further learning. Learning new examples requires higher precision updates or adaptable architecture~\cite{johnson2024edgeai}. This points to a need for algorithms that can learn under quantization constraints or dynamically increase precision when learning new information.
	
	Pruning is another model compression method that removes weights or neurons that are not essential, thereby creating a sparse model. Iterative pruning can drastically reduce model size with minimal loss in accuracy~\cite{han2015pruning}. In a continual learning context, pruning can serve a dual purpose: compressing the model and freeing capacity for new learning. Golkar et al.~\cite{golkar2019pruning} introduced a continual learning via neural pruning approach, which sparsifies the network after each task and makes the freed neurons available for learning subsequent tasks. They also introduce the notion of graceful forgetting, where a small amount of performance on past tasks is intentionally sacrificed in exchange for significant recovery of capacity to learn new tasks. This controlled trade-off is analogous to how the brain prunes synapses during development, eliminating some memory traces to optimize the network’s efficiency for future learning.
	
	Finally, sparsely activated architectures have gained attention for scaling AI models efficiently. A prime example is the Mixture-of-Experts (MoE) paradigm used in models like GLaM~\cite{du2022glam}, where a large number of sub-model “experts” are trained but only a small fraction are activated for any given input. This means the effective compute per sample remains modest even though the total parameter count is high, because most parameters stay inactive (dormant) per inference. Sparse activation is reminiscent of the brain’s sparse coding, i.e., only a subset of neurons fire for a given stimulus. In GLaM, this design achieved efficient scaling of language models, showing that one can increase model capacity (and thus potential knowledge) without proportional increases in runtime cost. For an edge scenario, one could envision a smaller-scale MoE where different experts specialize in different contexts or tasks (e.g., separate experts for different user activities or environments), and only the relevant expert is invoked at a time. This way, the device isn’t running an overly large model at once, and it could even turn off or prune experts that become irrelevant to the user’s life over time.
	
	Another promising approach for efficient inference is the use of early-exit networks, where models can terminate computation at intermediate layers for inputs deemed ``easy''~\cite{kaya2023earlyexit}. This dynamic inference strategy reduces average computational cost while retaining the ability to use deeper layers for more complex inputs. Such architectures are especially effective for edge deployment, enabling fast, energy-efficient responses for routine tasks while preserving full model depth when needed.
	
	In summary, the literature suggests that achieving continuous learning on the edge will require a combination of techniques: strategies to prevent forgetting, mechanisms inspired by the brain’s efficiency, and rigorous model compression. Next, we discuss specific neuroscience principles in detail and how they can contribute to a human-like continuous learning system.

	\section{Neuroscience Principles for Human-Like Learning in AI}
	The human brain exhibits several properties that are key to its ability to learn continually and efficiently. Here we highlight four such principles: Synaptic Pruning, Hebbian Plasticity, Sparse Coding, and Dual Memory Systems, describing their biological role and their implications for AI.

	\subsection{Synaptic Pruning}
\begin{figure}[htbp]
		\centering
		\includegraphics[width=0.9\linewidth]{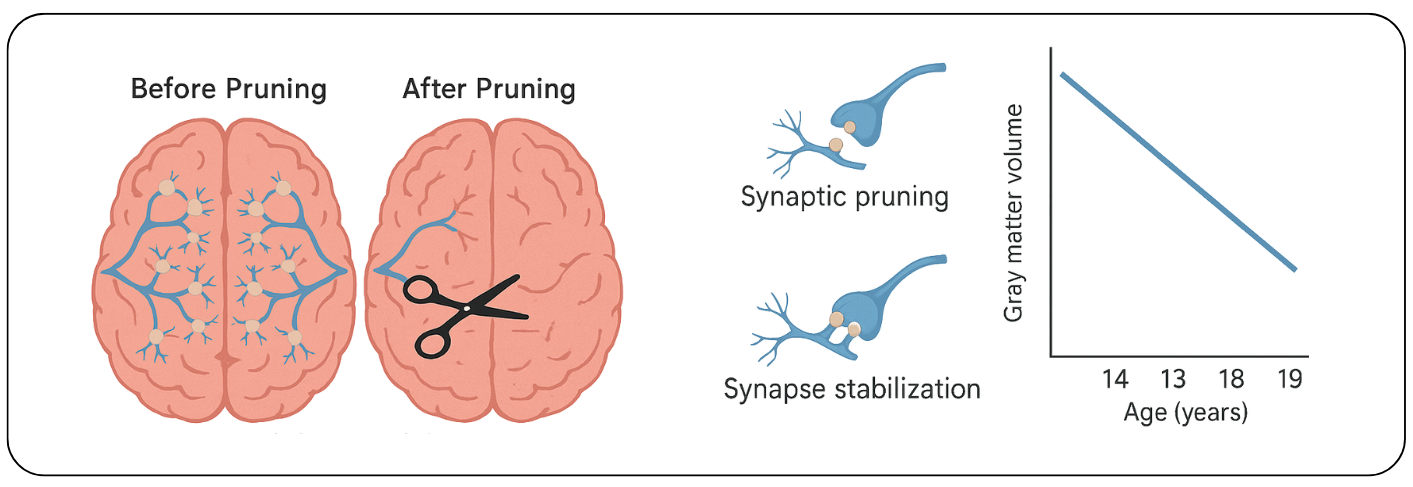}
		\caption{Illustration of synaptic pruning in the biological brain.}
		\label{fig:pruning-bio}
	\end{figure}

	During development and learning, brains not only form new synaptic connections but also eliminate a substantial number of existing synapses. Synaptic pruning is a process by which unused or weaker connections are removed, resulting in a more efficient and specialized network. In humans, synaptic pruning is especially prominent in early childhood and adolescence; it is estimated that roughly 50\% of synapses in some brain regions are pruned during adolescence~\cite{xie2023comorbidity}. By removing synapses that are no longer frequently used, the brain reduces metabolic cost and noise, effectively “decluttering” neural circuits~\cite{cafasso_pruning}. Pruning ensures that critical connections (those repeatedly used) are preserved and strengthened, while redundant pathways are cleared out (see Figure~\ref{fig:pruning-bio}).
	
\begin{figure}[htbp]
		\centering
		\includegraphics[width=0.9\linewidth]{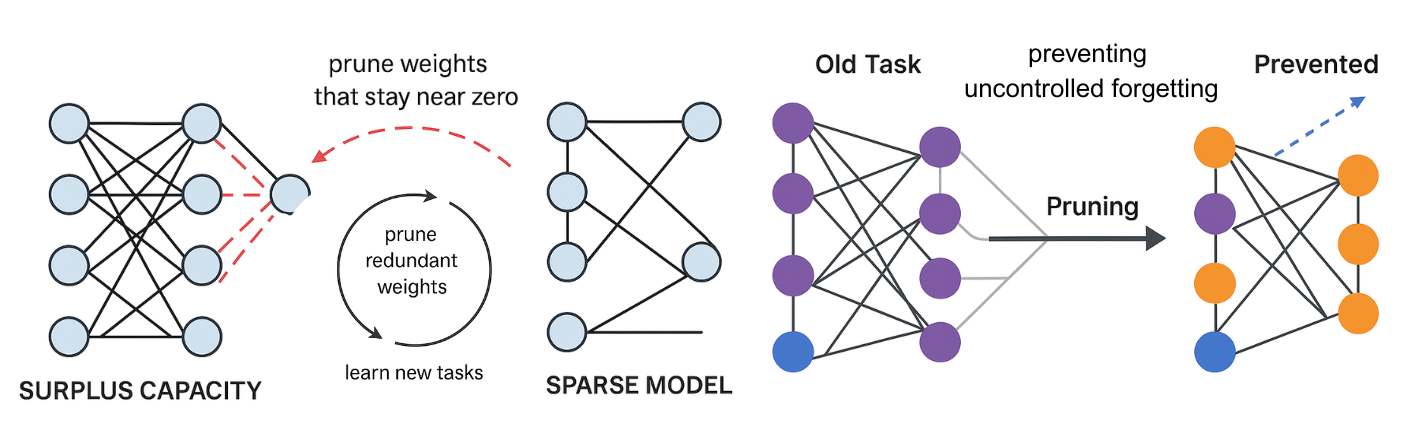}
		\caption{Synaptic pruning applied to AI.}
		\label{fig:pruning-ai}
	\end{figure}

	In the context of AI, synaptic pruning inspires methods to simplify neural networks during training or continual learning. A network can start with surplus capacity and then prune weights that have little influence on outputs (e.g., weights that stay near zero). This yields a sparse model that can be faster and more memory-efficient, which is crucial for edge devices. More importantly, in continual learning, pruning can reclaim capacity for new tasks. A prune-and-grow strategy can be employed: after learning a set of tasks, prune the redundant or least important weights, and then use the freed-up capacity (or even add new neurons if needed) to learn subsequent tasks. This mirrors biological neural development, where an initial overabundance of connections is later optimized (see Figure~\ref{fig:pruning-ai}).
	
	As noted earlier, Golkar et al.~\cite{golkar2019pruning} demonstrated that allowing a small amount of controlled forgetting of old tasks (through pruning) can prevent much larger uncontrolled forgetting during future training. Essentially, by pruning weights that encoded some old task details, the model accepts a minor performance hit on that task but gains flexibility to master new tasks without overlapping with old weights.
	
	\subsection{Hebbian Plasticity}
\begin{figure}[htbp]
		\centering
		\includegraphics[width=0.8\linewidth]{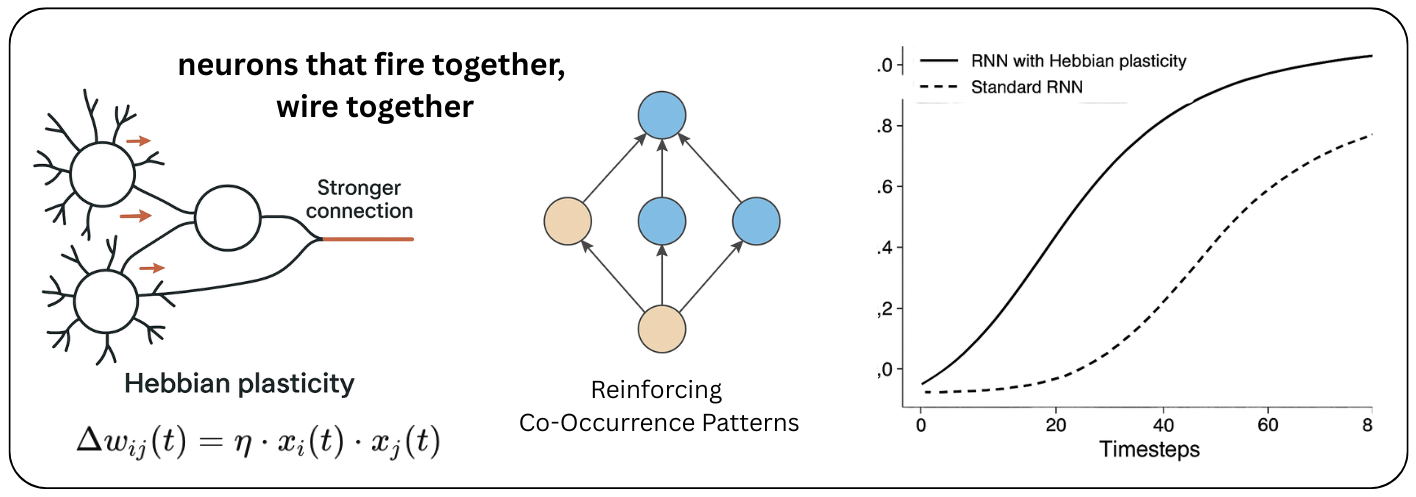}
		\caption{Donald Hebb's Principle.}
		\label{fig:hebb-basic}
	\end{figure}

	Donald Hebb’s famous principle, often paraphrased as “neurons that fire together, wire together”, captures the idea that the coincidence of activity in connected neurons tends to strengthen their connection (see Figure~\ref{fig:hebb-basic}). This Hebbian plasticity is considered a fundamental mechanism for associative learning: how we learn that two events or stimuli are related.
	
\begin{figure}[htbp]
		\centering
		\includegraphics[width=0.8\linewidth]{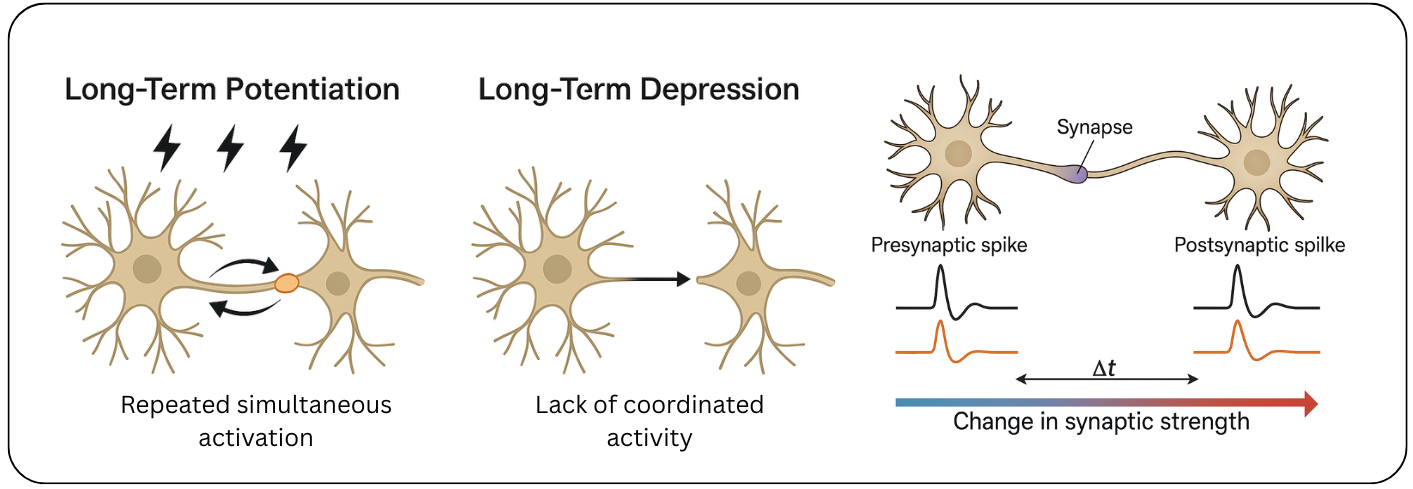}
		\caption{Spike-Timing-Dependent Plasticity (STDP).}
		\label{fig:hebb-stdp}
	\end{figure}

	In biological neurons, repeated simultaneous activation can lead to increases in synaptic weight (long-term potentiation), whereas lack of coordinated activity can lead to weakening (long-term depression). Variants of Hebbian learning rules, such as Spike-Timing-Dependent Plasticity (STDP), adjust synaptic strengths based on the precise timing of spikes between neurons (see Figure~\ref{fig:hebb-stdp}).
	
	Translating Hebbian learning to AI offers a way for networks to learn without explicit labels or gradient-based training, by simply reinforcing co-occurrence patterns. A direct emulation is seen in unsupervised feature learning methods like Hebbian neural networks and competitive learning, which were historically explored. For example, Oja’s rule for principal component extraction is a modified form of Hebbian learning~\cite{oja1982hebb}.
	
\begin{figure}[htbp]
		\centering
		\includegraphics[width=0.8\linewidth]{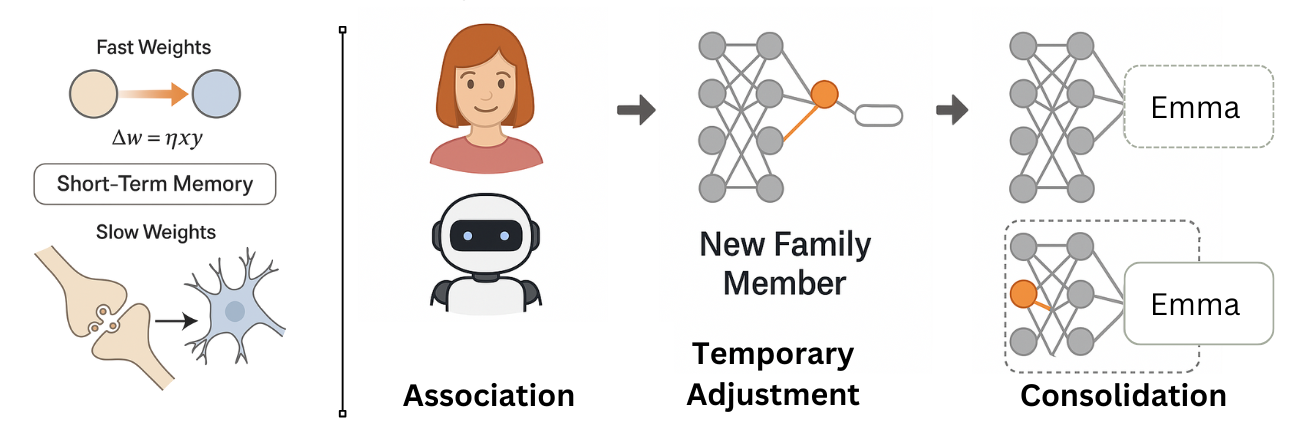}
		\caption{Hebbian Plasticity and Rapid Personalization.}
		\label{fig:hebb-fastweights}
	\end{figure}

	In modern contexts, Hebbian principles resurface in fast adaptation components. For instance, one can design certain weights in a network to update according to a simple rule:
	
	\[
	\Delta w = \eta \cdot x \cdot y
	\]
	
	where \( x \) and \( y \) are the pre- and postsynaptic activations, respectively, and \( \eta \) is a learning rate. This formulation allows the network to form temporary memories on the fly (see Figure~\ref{fig:hebb-fastweights}).
	
	Miconi et al.~\cite{miconi2018plasticity} incorporated differentiable Hebbian plasticity into recurrent neural networks, enabling the model to adjust specific synapses quickly based on activity during a task. They demonstrated that such models perform better on sequential learning benchmarks.
	
	Likewise, the concept of fast weights~\cite{ba2016fastweights} uses Hebbian updates to temporarily store information that a standard slow-weight network can later retrieve. This mechanism is analogous to short-term memory at the synaptic level, with Hebbian plasticity serving as a biological metaphor for rapid within-task adaptation in artificial systems.

	\subsection{Sparse Coding}
\begin{figure}[htbp]
		\centering
		\includegraphics[width=0.85\linewidth]{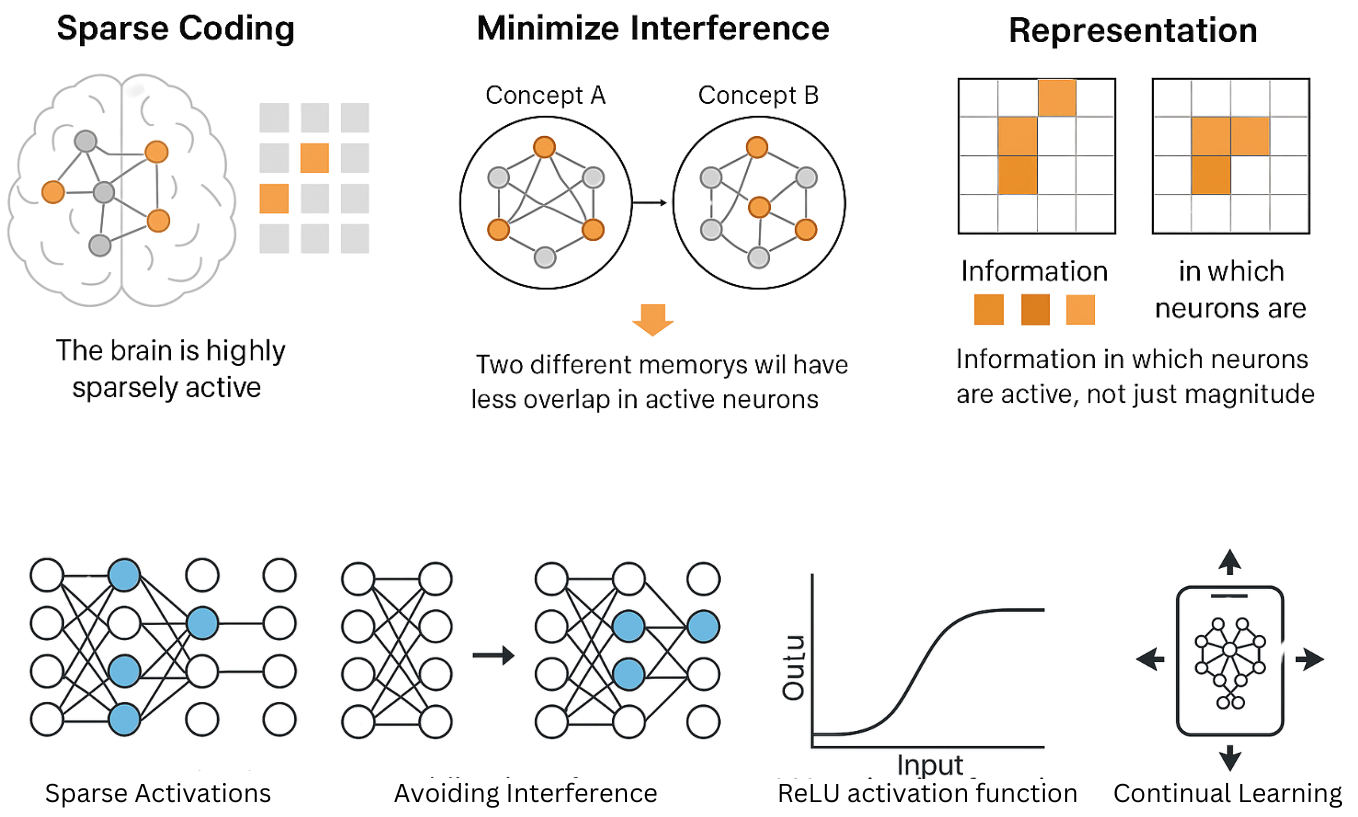}
		\caption{Sparse Coding improves efficiency, reduces overlap}
		\label{fig:sparse-neuro}
	\end{figure}

	The brain is highly sparsely active: at any given moment, only a small fraction of neurons in a region fire strongly, while most remain relatively quiet. This sparse coding improves efficiency and reduces overlap between representations. By activating a unique (sparse) combination of neurons for each concept or stimulus, the brain minimizes interference, i.e., two different memories will have less overlap in active neurons, reducing the chance they disrupt each other. Sparse representations also carry information in which neurons are active, not just the magnitude, increasing the representational capacity for a given number of neurons (see Figure~\ref{fig:sparse-neuro}). Empirical studies in neuroscience, such as recordings from the visual and auditory cortices, have found evidence of sparse firing patterns, e.g., a given cortical neuron might respond strongly only to a very specific stimulus feature and remain dormant otherwise. This supports Barlow’s efficient coding hypothesis that the brain aims to represent information with minimal redundancy.
	
	In AI, inducing sparse coding in neural networks can be beneficial for similar reasons. Techniques like L1 regularization on activations, winner-take-all layers, or the use of activation functions that promote sparsity (e.g., ReLU inherently zeroes out negative inputs) lead to networks where for any input, many neurons are exactly zero or near inactive. Sparse neural networks not only yield computational speed-ups (since many units do nothing for a given input) but also reduce interference between tasks in continual learning. If different tasks or classes activate different subsets of neurons, the overlap in weights usage is lower, hence less catastrophic forgetting.
	
\begin{figure}[htbp]
		\centering
		\includegraphics[width=0.85\linewidth]{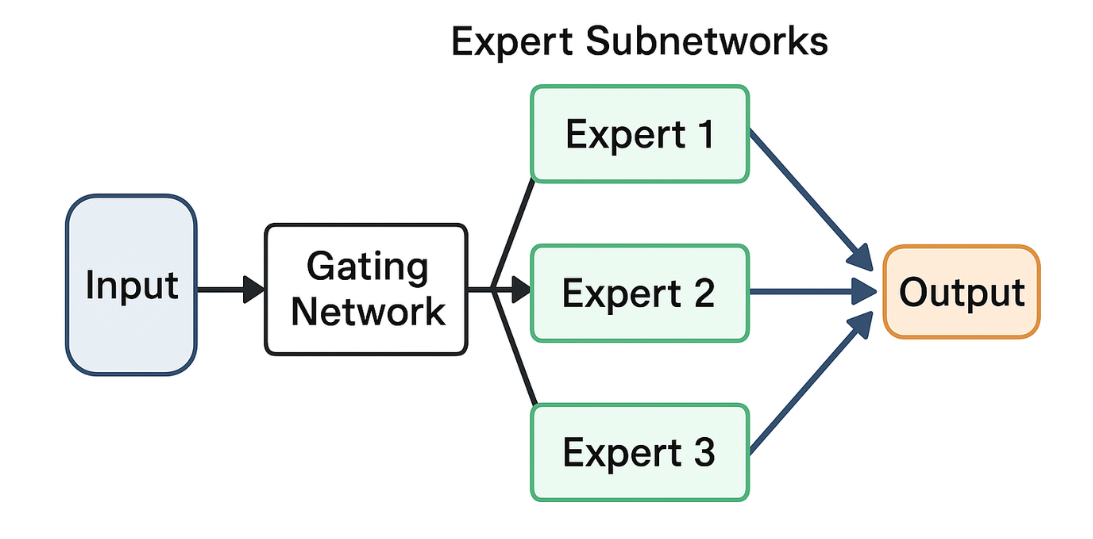}
		\caption{Mixture-of-Experts Model}
		\label{fig:sparse-ai}
	\end{figure}

	Recent continual learning research leverages this. SparCL~\cite{wang2022sparcl} explicitly encouraged a sparse subset of features to specialize for each task and showed improved retention when training on an edge device with limited memory. Moreover, the previously mentioned mixture-of-experts model (GLaM) is an example of architected sparsity at a coarse level. Only a few expert subnetworks out of many are active for any input~\cite{du2022glam} (see Figure~\ref{fig:sparse-ai}).
	
	One can imagine a continual learning system that develops separate “expert modules” for different contexts the agent encounters (e.g., work vs. home, or indoor vs. outdoor for a robot), and a gating mechanism (possibly learned or context-driven) ensures only the relevant ones activate. This way, knowledge is compartmentalized, and adding a new context would mean recruiting a new sparse set of neurons (or an expert) without disturbing the others.
	
	\subsection{Dual Memory Systems (Fast and Slow Learning)}
\begin{figure}[htbp]
		\centering
		\includegraphics[width=0.85\linewidth]{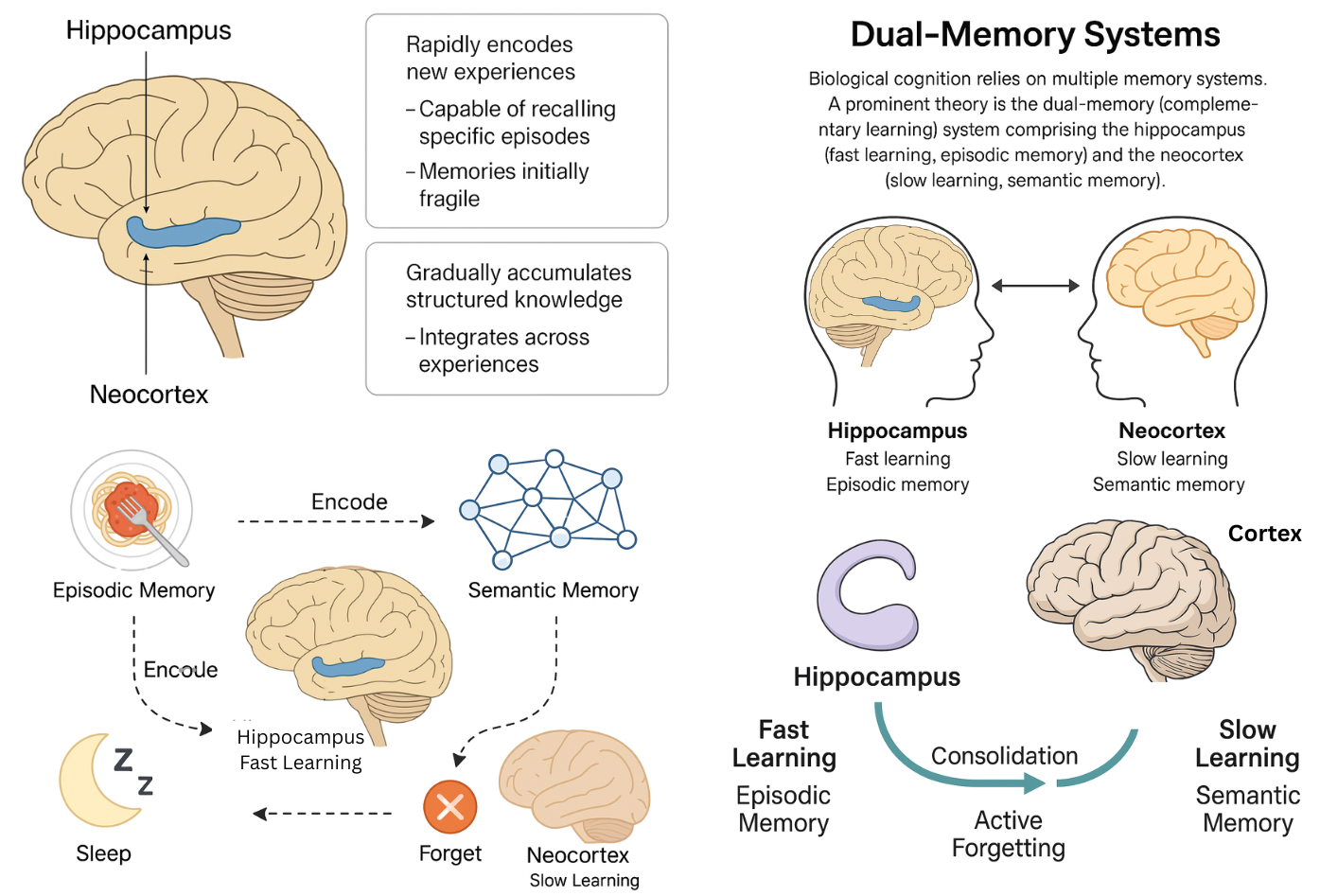}
		\caption{Fast and Slow Learning - hippocampus vs neocortex}
		\label{fig:dual-memory-bio}
	\end{figure}

	Biological cognition relies on multiple memory systems. A prominent theory is the dual-memory (complementary learning) system comprising the hippocampus (fast learning, episodic memory) and the neocortex (slow learning, semantic memory). The hippocampus can rapidly encode new experiences (within one or a few exposures) and is capable of recalling specific episodes (e.g., what you ate for dinner last night), but these memories are initially fragile. The cortex, on the other hand, gradually accumulates structured knowledge (like abstract concepts and skills) over repeated exposures and sleep cycles, integrating across experiences (see Figure~\ref{fig:dual-memory-bio}).
	
	Over time, memories can be consolidated from hippocampus to cortex: important events are replayed and gradually incorporated into the cortical networks for long-term storage, allowing the hippocampus to remain available for new memories. This division allows humans to both learn quickly (via hippocampal plasticity) and retain stability and generalization (via cortical slow learning)~\cite{mcclelland2020integration}. Notably, active forgetting is part of this process. The hippocampus discards or weakens memories that are not revisited or useful, preventing overload.
	
\begin{figure}[htbp]
		\centering
		\includegraphics[width=0.85\linewidth]{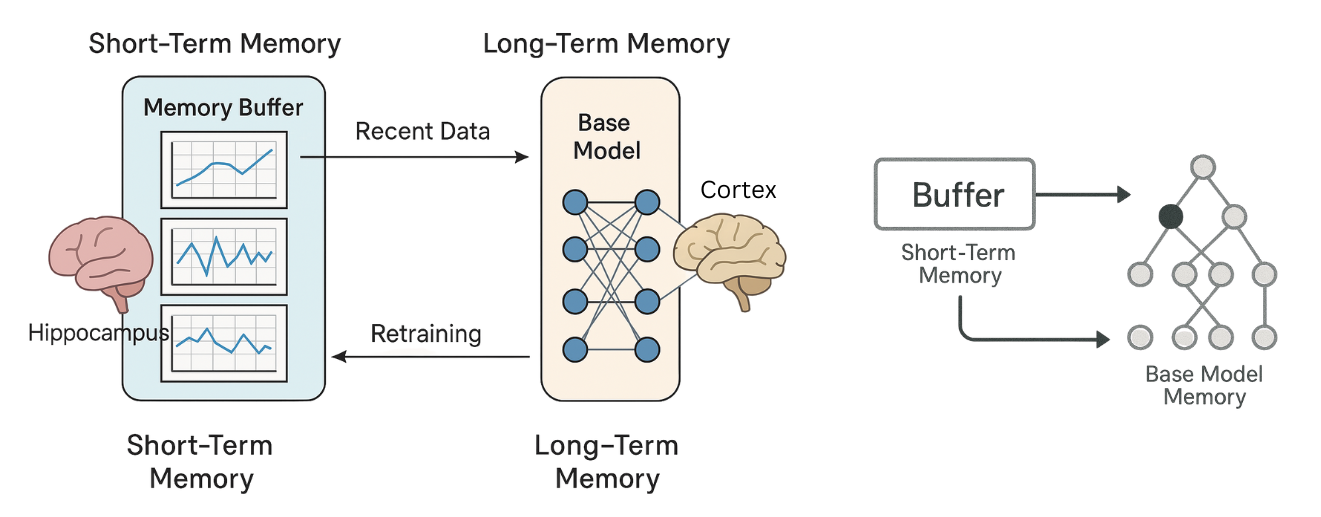}
		\caption{Short term and long-term memory in AI}
		\label{fig:dual-memory-ai}
	\end{figure}

	For AI, implementing a dual memory system can address the stability–plasticity dilemma by design. Many researchers have proposed complementary learning architectures. One simple instantiation is to have a main model that learns gradually (e.g., via standard Stochastic Gradient Descent on accumulated knowledge) and an auxiliary memory module that can store recent experiences explicitly. For example, a memory buffer plus base model approach: the buffer (akin to hippocampus) caches recent data or micro-models of recent tasks, and the base model (akin to cortex) is intermittently retrained or expanded using information from the buffer. This idea underlies some replay-based methods where the buffer is the short-term memory and the network’s weights are the long-term memory (see Figure~\ref{fig:dual-memory-ai}).
	
	The Titans architecture~\cite{behrouz2025titans} offers a more structural approach, in which the attention mechanism acts as short-term contextual memory, while the long-term memory module serves as durable storage of knowledge that can be queried. Another example is meta-learning approaches like Memory-Augmented Neural Networks~\cite{santoro2016mann} or Online Aware Meta-Learning~\cite{javed2019online}, which effectively train a model to have two parts: one that updates quickly (sometimes even at test time or deployment) and one that changes slowly (across meta-iterations). These are all reflections of the fast/slow learning split.
	
	The dual memory concept directly tackles catastrophic forgetting: the long-term memory is guarded against rapid changes, so it maintains past knowledge, while the short-term memory handles quick learning of new things without immediately overwriting the long-term parameters. Over time, only vetted changes make it into a long-term store. There are challenges, such as determining how large each memory should be on an edge device and ensuring consistency between the two memories so that the AI’s behavior remains coherent. However, many continual learning frameworks implicitly or explicitly implement this idea, showing its promise. For instance, “Learn to Grow”~\cite{li2019learntogrow} uses a knowledge base network (slow learning) and an adaptive expansion network for new tasks (fast learning) that later merges with the base. Similarly, in reinforcement learning, we can separate a policy into a stable pre-trained part and an adaptable online part that learns new tricks quickly in a new environment.
	
	The principles discussed above not only highlight core mechanisms behind human learning but also provide a scaffold for developing AGI systems with similar capabilities. In the next section, we describe how these principles are operationalized into a coherent architecture.

	\section{Proposed Architecture for Personalized AGI on the Edge}
	
	Drawing on the insights above, we propose an AI architecture with a \textbf{Tri-Memory Continual Learning} system tailored for deployment on resource-constrained edge devices, such as humanoid robots powered by an NVIDIA Jetson platform. The overarching goal is to realize concurrent inference and real-time training while minimizing computational overhead. By integrating Hebbian-like learning rules with error-driven backpropagation, the model continuously adapts to new information, prunes rarely used pathways, and maintains essential knowledge across short-term, long-term, and permanent memory modules. This design not only aligns with the biological principles of selective forgetting and retention but also offers a practical pathway toward personalized AGI, where an embodied agent can evolve and specialize its cognitive functions in real-world environments.
	
	\subsection{Motivations and Key Concepts}
	
	Modern humanoids, robots, and other edge devices must process continuous data streams in real time, despite having only modest on-board computing and limited access to off-device resources. Drawing on principles of Hebbian plasticity and Dual-Memory Theory, our approach is designed to enable continuous learning with selective forgetting, much like synaptic adjustments observed in the human brain.
	
	Instead of performing large offline training cycles, the system implements immediate, lightweight updates after each inference through a local usage tracking mechanism, and relies on periodic background processes for heavier consolidation activities. This synergy of rapid, incremental plasticity with delayed, more substantial network restructuring aligns with neuroscience notions of fast versus slow learning.
	
	A central element of our method is the use of \textbf{microsleeps}. These are brief intervals on the order of milliseconds to a few seconds, during which forward inference is momentarily paused or deprioritized. This microsleep window allows the model to perform a minimal form of simulated synaptic decay, referred to here as a \textit{global offset}: a small uniform negative shift applied to every weight. Weights that remain above zero after the offset are maintained, while those dropping below zero effectively become inactive.
	
	However, these microsleeps do not include replay-based rehearsals or actual large-scale pruning. By limiting microsleeps to only performing the global offset, the system achieves a continuous, gentle decay mechanism without incurring the high computational costs of more invasive tasks like pruning or full retraining.
	
	Actual synaptic pruning in our framework occurs during longer “offline” or nightly sessions when the robot can access more plentiful power and is less constrained by real-time operational demands. During these sessions, the pruning threshold is \textbf{adaptively determined} based on daily usage statistics, echoing the biological process by which neural circuits eliminate underutilized synaptic connections over time.
	
	If, for instance, the robot encounters few novel stimuli on a given day, the system can determine that large-scale pruning is unnecessary. Conversely, a day of extensive novel experiences could trigger a more substantial reallocation of network resources. This strategic scheduling of pruning allows for dynamic sparse coding, in which rarely used weights are trimmed away and the overall parameter space remains compact and energy-efficient.
	
	In addition to pruning, replay-based rehearsals aimed at mitigating catastrophic forgetting are also shifted to these offline windows, drawing parallels with the idea of sleep-based consolidation observed in biological brains. A small batch of recent experiences, along with examples from a curated replay buffer, is revisited in a brief training pass. This ensures that newly formed associations are reinforced while critical long-standing knowledge remains intact.
	
	This is consistent with a dual-memory (or, in our expanded case, \textbf{tri-memory}) system: transient, plastic changes accumulate in the short-term store, but only through deliberate rehearsal and consolidation do robust, long-term representations form.
	
	Overall, our architecture combines neuroscientific insights, like Hebbian-like local updates for immediate adaptation, synaptic pruning for resource management, microsleep offsets for lightweight continuous decay, and offline replay for consolidating indispensable skills, to sustain robust yet efficient continual learning on the edge. By deferring more costly processes (e.g., pruning, replay training) to offline or nightly sessions, the system remains agile and power-thrifty during active hours, fostering the development of \textbf{Personalized AGI} in embodied devices with stringent resource constraints.

	\subsection{Tri-Memory System: STM, LTM, and PM}
	
\begin{figure}[htbp]
		\centering
		\includegraphics[width=0.85\linewidth]{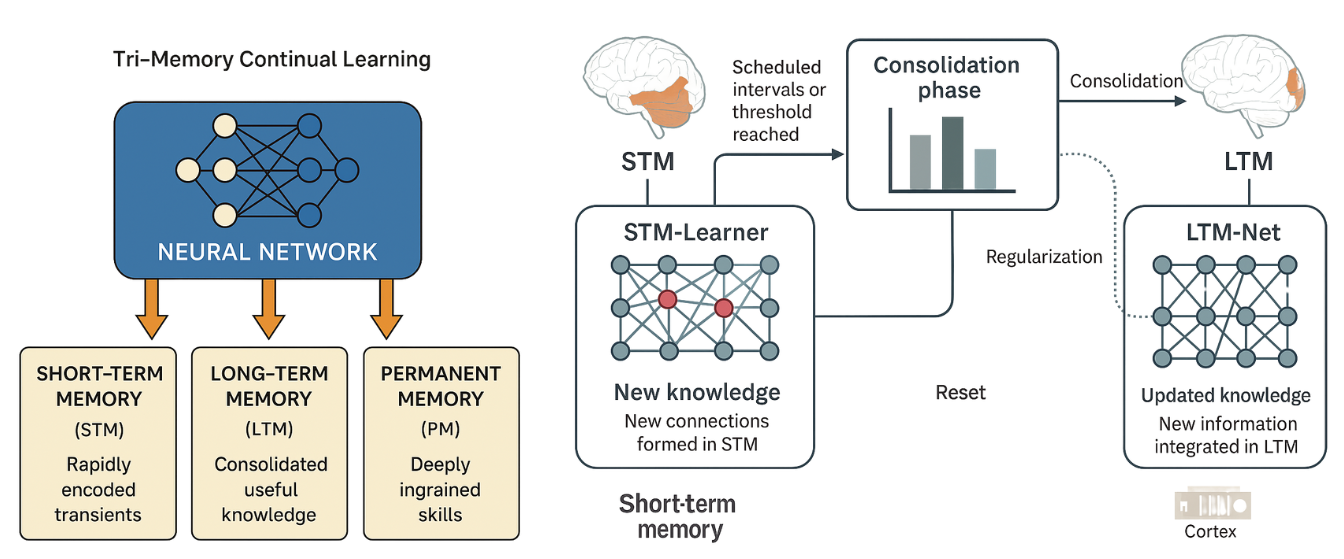}
		\caption{Tri-Memory System: STM, LTM, and PM memory modules with dynamic consolidation}
		\label{fig:tri-memory}
	\end{figure}

	A central component of our approach is the \textbf{Tri-Memory Continual Learning} design, which extends beyond the familiar dual-memory framework by introducing a dedicated \textit{Permanent Memory (PM)} module (see Figure~\ref{fig:tri-memory}). This architecture is inspired by neuroscience concepts in which memory formation and consolidation occur at multiple timescales, reflecting a spectrum from rapidly encoded transients to deeply ingrained skills. By partitioning the model’s parameters across three distinct tiers: \textbf{Short-Term Memory (STM)}, \textbf{Long-Term Memory (LTM)}, and \textbf{Permanent Memory (PM)}, the system aligns with the brain’s capacity to capture new experiences quickly, consolidate what proves repeatedly useful, and permanently preserve core competencies.
	
	The \textbf{Short-Term Memory (STM)} serves as a fast-learning buffer for novel information. When the robot encounters unanticipated inputs or stimuli, the STM’s plastic connections adapt rapidly through Hebbian-like updates, allowing the system to incorporate new correlations on the fly. This high degree of plasticity ensures swift responses to immediate demands; however, because the STM is prone to frequent fluctuations, knowledge stored here remains relatively volatile.
	
	In contrast, the \textbf{Long-Term Memory (LTM)} stabilizes useful patterns that have repeatedly demonstrated their relevance. Mechanisms such as usage tracking and minimal error-driven updates identify which STM connections deserve migration into the more enduring LTM. Once in LTM, parameters are less frequently modified, reducing the risk of overwriting well-established representations. Nonetheless, LTM weights can still be pruned or decayed during offline maintenance sessions if they fall below adaptive thresholds, reflecting the ongoing need to balance retention with resource efficiency.
	
	\textbf{Consolidation Phase (periodic):} When STM’s accumulated changes reach a threshold or at scheduled intervals, the system enters a consolidation phase. The new knowledge in STM is analyzed: if some of it should become long-term (e.g., frequently used skills or significant improvements on a task), the \textit{Consolidation Module} updates the LTM-Net. This could involve a brief training session where the LTM-Net is fine-tuned on a combined dataset of its original core knowledge plus a sample of new data recorded by STM. Regularization methods (such as EWC) are applied to ensure important weights don’t change drastically. Once integrated, the LTM-Net includes the new knowledge in a stable form. After consolidation, the STM-Learner may be reset or cleared of the consolidated items, similar to hippocampal-to-cortex transfer freeing up hippocampus.
	
	The third tier, \textbf{Permanent Memory (PM)}, captures a model’s most indispensable features or abilities, safeguarding them from routine decay and pruning. Only after a parameter or sub-network consistently demonstrates high utility does it graduate from LTM to PM, ensuring that mission-critical functionalities, skills, and memories are preserved indefinitely. Inspired by the long-lasting or ``hardwired'' skills in biological systems, PM focuses on core capabilities deemed too crucial to risk losing, such as fundamental locomotion or essential perceptual routines in humanoid robots. By incorporating this extra layer of protected knowledge, the Tri-Memory design facilitates continuous adaptation without jeopardizing hard-won competencies that form the backbone of a robot’s broader intelligence.
	
\begin{figure}[htbp]
		\centering
		\includegraphics[width=0.85\linewidth]{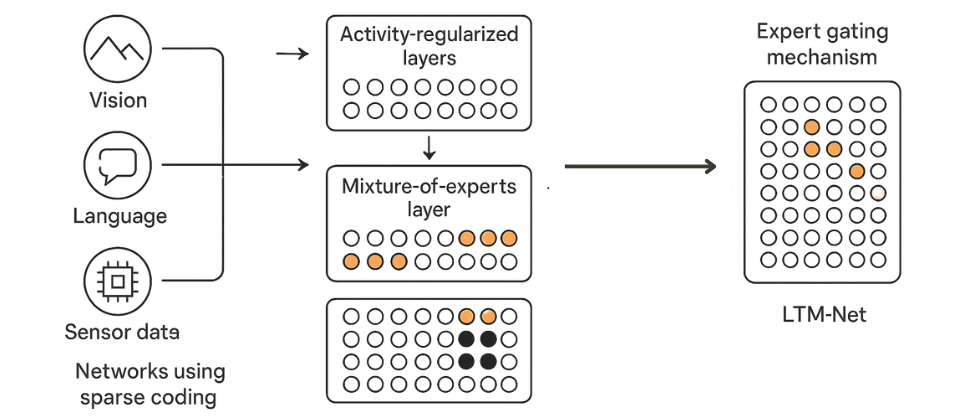}
		\caption{Sparse distributed representation}
		\label{fig:moe-sparse}
	\end{figure}

	\textbf{Sparse Distributed Representation:} Each network uses sparse coding principles so that any given input (or task) activates only a subset of neurons. This is achieved by using layers with activity regularization (encouraging most neurons to be inactive) or architecturally via a mixture-of-experts (MoE) layer (see Figure~\ref{fig:moe-sparse}). Concretely, in LTM-Net, we incorporate an expert gating mechanism: for different domains of knowledge (e.g., vision, language, or other contexts), only the corresponding expert sub-network is activated. If a new domain of data is encountered (e.g., the mobile assistant begins processing a new sensor type), a new expert can be added rather than interfering with existing ones.
	
	This echoes the MoE idea from GLaM~\cite{du2022glam}, but adapted for resource-constrained edge devices. The gating could use context features to choose which expert to use for a given input, ensuring modularity and preserving previous capabilities.

	\subsection{Microsleeps for Lightweight Decay}
	
\begin{figure}[htbp]
		\centering
		\includegraphics[width=0.85\linewidth]{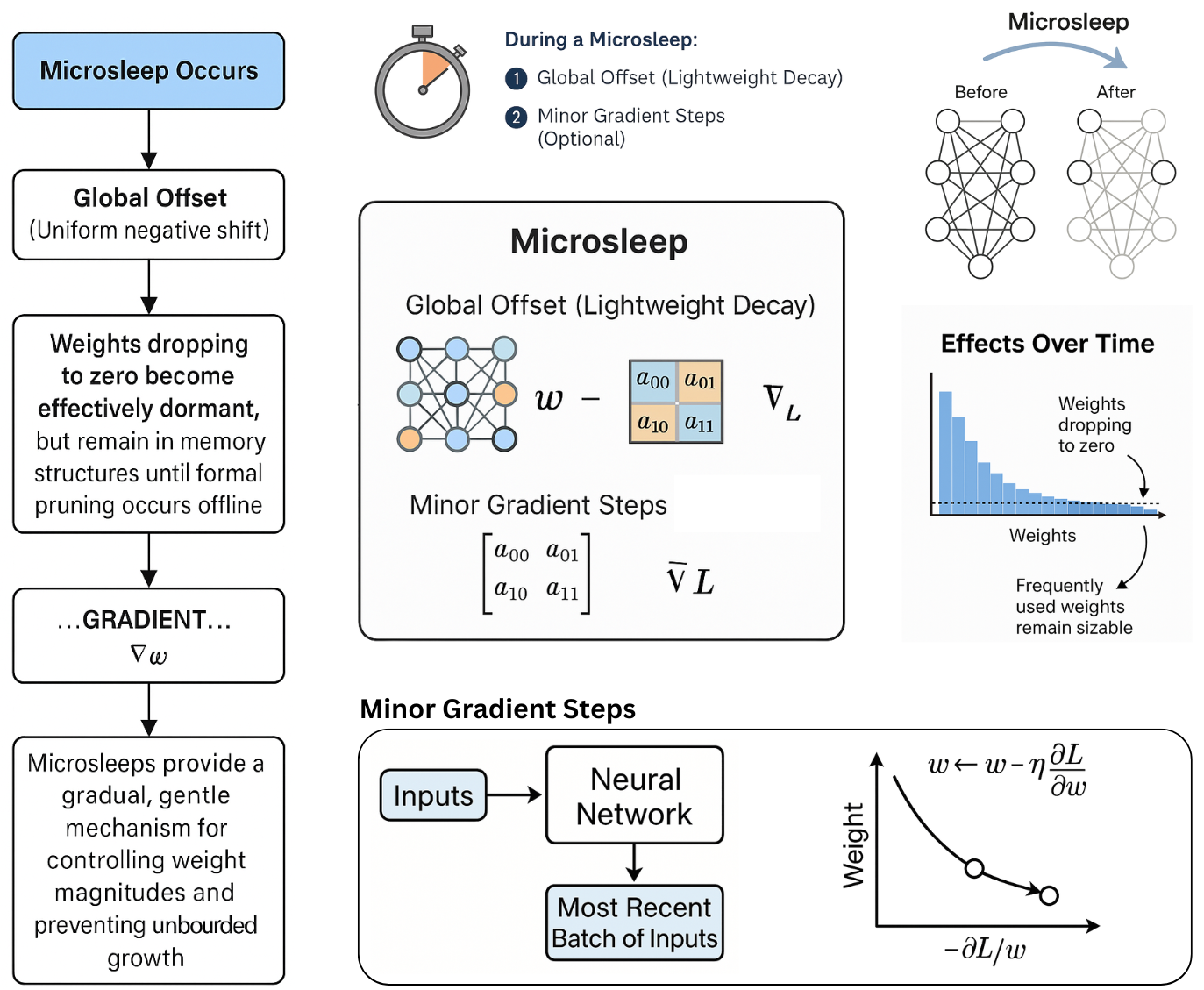}
		\caption{Microsleeps for Lightweight Decay: Global Offset and Minor Gradient Steps}
		\label{fig:microsleep-decay}
	\end{figure}

	In biologically inspired frameworks, short rest periods, \textit{microsleeps}, offer opportunities to enact small but beneficial housekeeping operations. In our implementation, microsleeps occur at periodic intervals or after a set number of inferences; each interval typically spans milliseconds to seconds, depending on the application’s latency tolerance (see Figure~\ref{fig:microsleep-decay}).
	
	During a microsleep:

	\textbf{Global Offset (Lightweight Decay):} A uniform negative shift is applied to all weights in the network. This simulates synaptic decay but in a single, vectorized operation that is computationally efficient. Weights dropping to zero become effectively dormant but remain in memory structures until formal pruning occurs offline (Section~\ref{sec:nightly-pruning}).
	
	\textbf{Minor Gradient Steps (Optional):} If system resources allow, a brief gradient-based update can be performed on the most recent batch of inputs. However, microsleeps are intentionally designed to be short, minimizing overhead so that real-time inference is not substantially disrupted.

	Large-scale pruning is \textbf{not} performed during these micro-sleeps. Instead, microsleeps provide a gradual, gentle mechanism for controlling weight magnitudes and preventing unbounded growth. The immediate impact on overall memory footprint is small, but cumulatively, global offsets nudge the network toward a stable equilibrium in which frequently used weights remain sizable, and rarely used weights decay toward negligible values.
	
	\subsection{Nightly (Offline) Pruning and Replay-Based Training}
	\label{sec:nightly-pruning}
	
	\begin{figure}[htbp]
		\centering
		\includegraphics[width=1.0\linewidth]{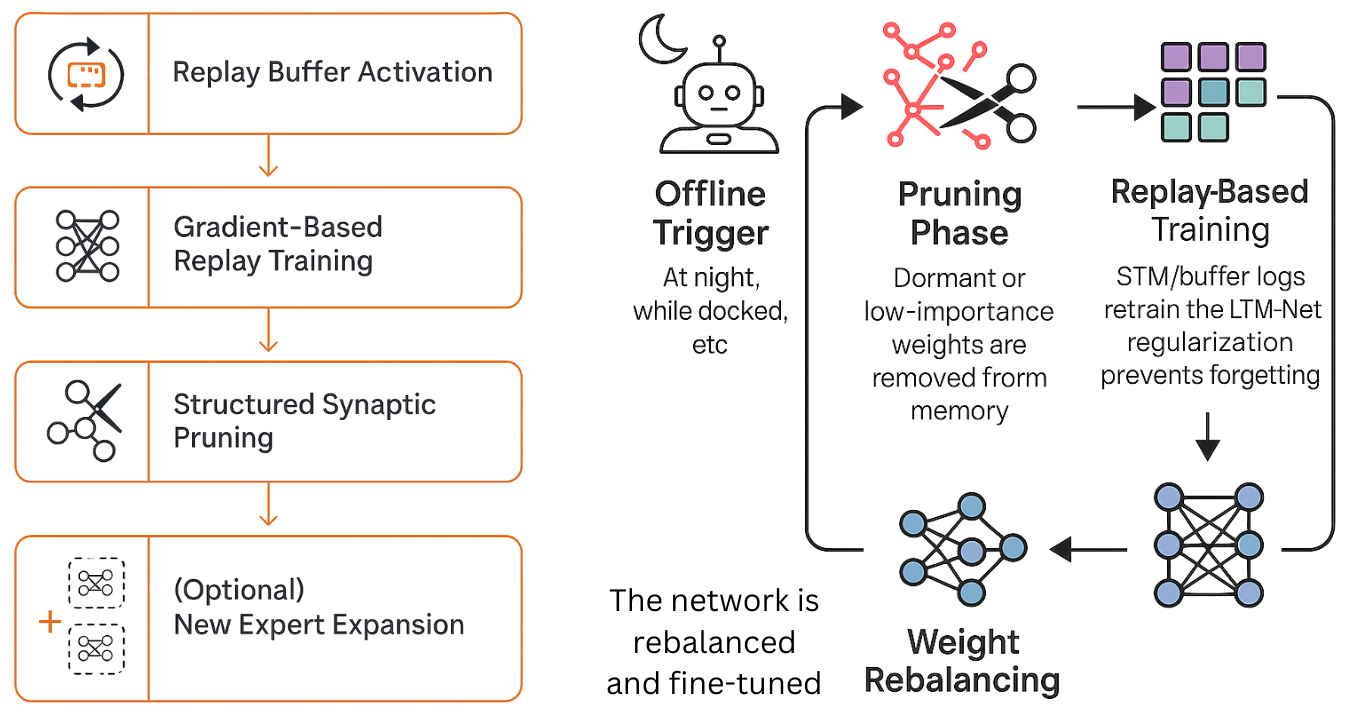}
		\caption{Nightly (Offline) Pruning}
		\label{fig:nightly-offline}
	\end{figure}
	
	To handle more intensive tasks such as synaptic pruning and catastrophic forgetting avoidance, we reserve \textbf{offline sessions}, often referred to in neuroscience as analogs of ``sleep-based consolidation.'' During these extended sessions, the robot (or edge device) is typically docked or otherwise connected to a stable power source, relaxing the stringent energy constraints of real-time operation (see Figure~\ref{fig:nightly-offline}).
	
	In this offline phase, the model applies \textbf{adaptive pruning}, a process guided by usage statistics accumulated throughout the day. Each connection, whether a neuron, attention head, or weight, maintains counters that record how often it contributes meaningfully to successful inferences. If these counters remain persistently low, the connection is deemed superfluous. The decision to prune is based on dynamically determined thresholds that factor in the robot’s memory capacity and the amount of fresh information acquired since the last offline cycle.
	
\begin{figure}[htbp]
		\centering
		\includegraphics[width=1.0\linewidth]{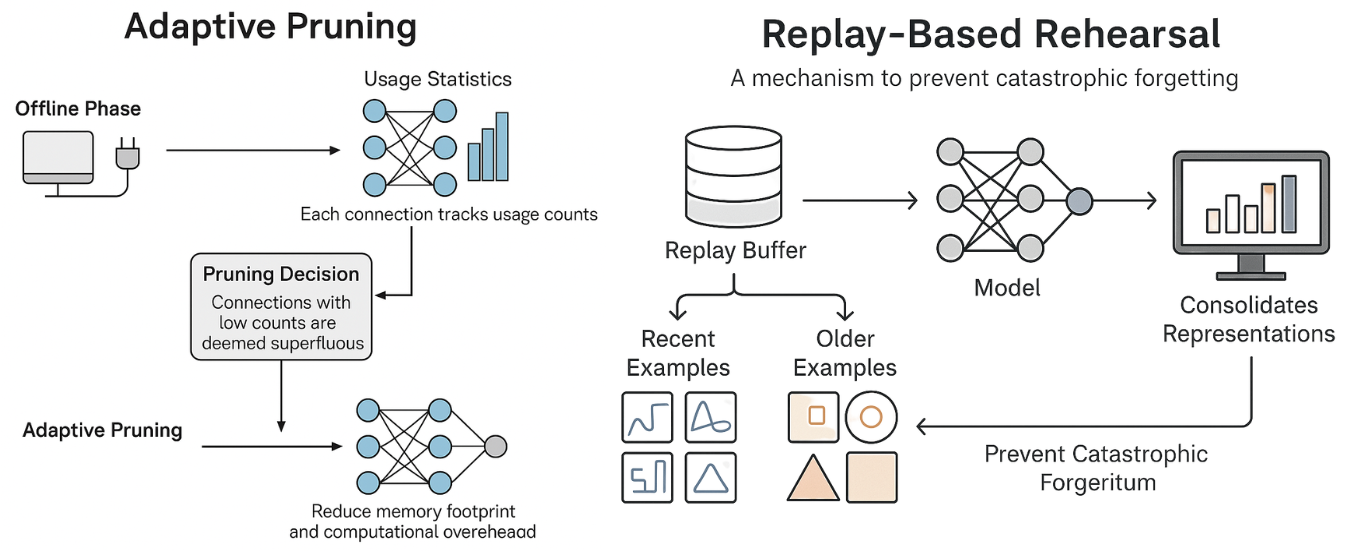}
		\caption{Adaptive Pruning and Replay-Based Rehearsal}
		\label{fig:nightly-pruning}
	\end{figure}

	By pruning underused connections, the system achieves a sparser internal representation, reducing both memory footprint and computational overhead for subsequent online operations. Importantly, this process is \textit{not} forced to run every night. In cases where there has been minimal novel interaction or data, the system may skip pruning, preserving established parameters and preventing undue knowledge loss (see Figure~\ref{fig:nightly-pruning}).
	
	While usage counters provide a quantitative basis for pruning and consolidation, they do not fully capture the \textbf{qualitative} aspect of interactions. To address this, we propose integrating lightweight user feedback mechanisms. After each interaction session, the system optionally solicits an explicit rating (e.g., a 1–5 scale) or applies a small sentiment analysis model to infer feedback implicitly. This sentiment signal acts as a \textit{modulating factor} during consolidation.
	
	For example, a frequently used connection receiving consistently negative sentiment can be deprioritized or tagged for intentional replacement, even if usage is high. Conversely, a rarely used skill that receives strong positive sentiment could be retained longer or reinforced during replay. Over time, persistently negative feedback on a consolidated skill (in LTM or PM) can trigger its replacement with an updated version fetched from trusted external sources.
	
	The offline window also includes \textbf{replay-based rehearsal}, a mechanism to prevent catastrophic forgetting. Over the course of normal operation, the model accumulates a small replay buffer containing examples that capture recently encountered patterns, as well as older, foundational samples critical to the robot’s performance. During offline sessions, the system re-presents these examples in a brief, small-batch training pass.
	
	This rehearsal process consolidates newly formed representations while reinforcing established competencies, ensuring that fresh updates do not overwrite previously mastered knowledge. By periodically revisiting past tasks in tandem with new experiences, the architecture maintains robust performance across a range of scenarios, aligning with biological theories that memory consolidation is facilitated by repeated exposure to key stimuli outside of direct task execution.
	
	Through these processes, the model balances \textbf{adaptivity with stability}. The offline pruning step enforces sparse coding, a neuroscientific principle that fosters computational efficiency, while replay-based rehearsal consolidates essential competencies into LTM or PM.

	\subsection{Hybrid Learning: Hebbian and Error-Driven Updates}
	
	In addition to determining \textit{what} knowledge is retained, the architecture also addresses \textit{how} it is encoded through hybrid learning mechanisms. To unify local correlation-based plasticity with global performance optimization, the system employs both \textbf{Hebbian-like} and \textbf{error-driven} weight adjustments.
	
	Whenever two neurons co-activate in a beneficial manner, as indicated by usage counters and partial credit assignment, their connecting weights in STM or LTM receive a modest increment. These Hebbian updates reinforce useful associations with minimal latency overhead, promoting the growth of correlations in a biologically plausible manner.
	
\begin{figure}[htbp]
		\centering
		\includegraphics[width=0.85\linewidth]{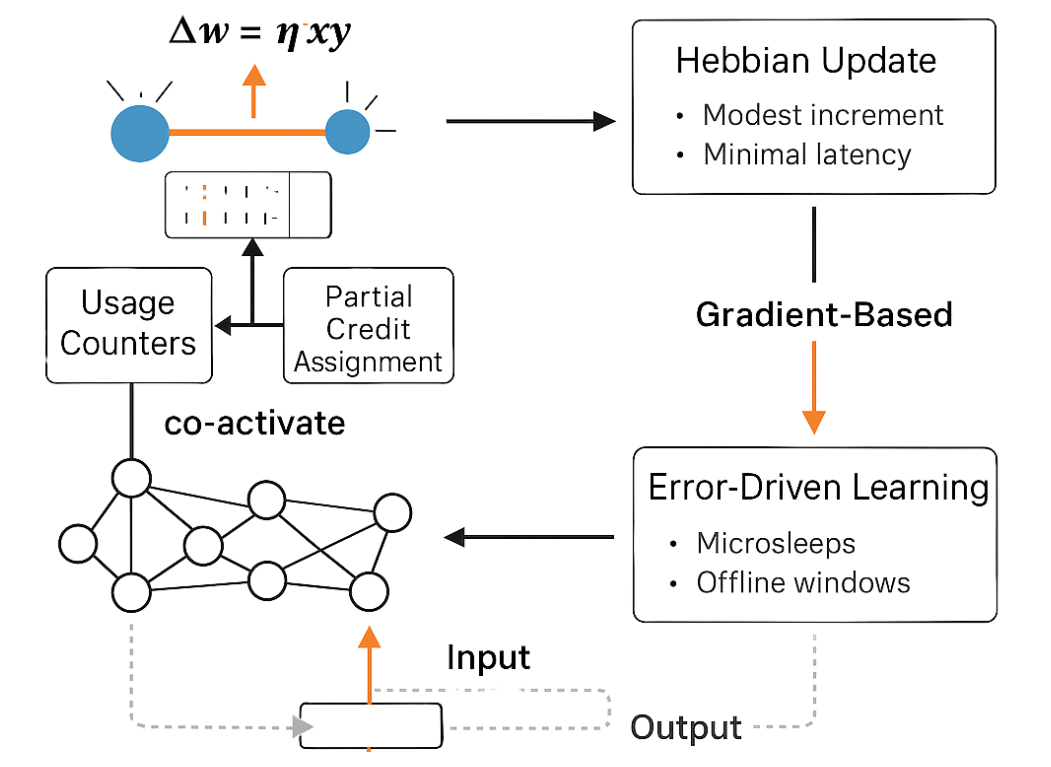}
		\caption{Hybrid learning with Hebbian and gradient-based updates occurring at different phases of the system's timeline.}
		\label{fig:hybrid-learning}
	\end{figure}

	In parallel, \textbf{error-driven (gradient-based)} learning is applied selectively during microsleeps (for negligible micro-corrections) or more extensively in offline windows (for thorough fine-tuning). This approach complements Hebbian plasticity by aligning the network with task objectives, effectively integrating local correlation signals with global performance requirements (see Figure~\ref{fig:hybrid-learning}).
	
	\subsection{Timeline of Operations}
	
	\begin{figure}[htbp]
		\centering
		\includegraphics[width=1.0\linewidth]{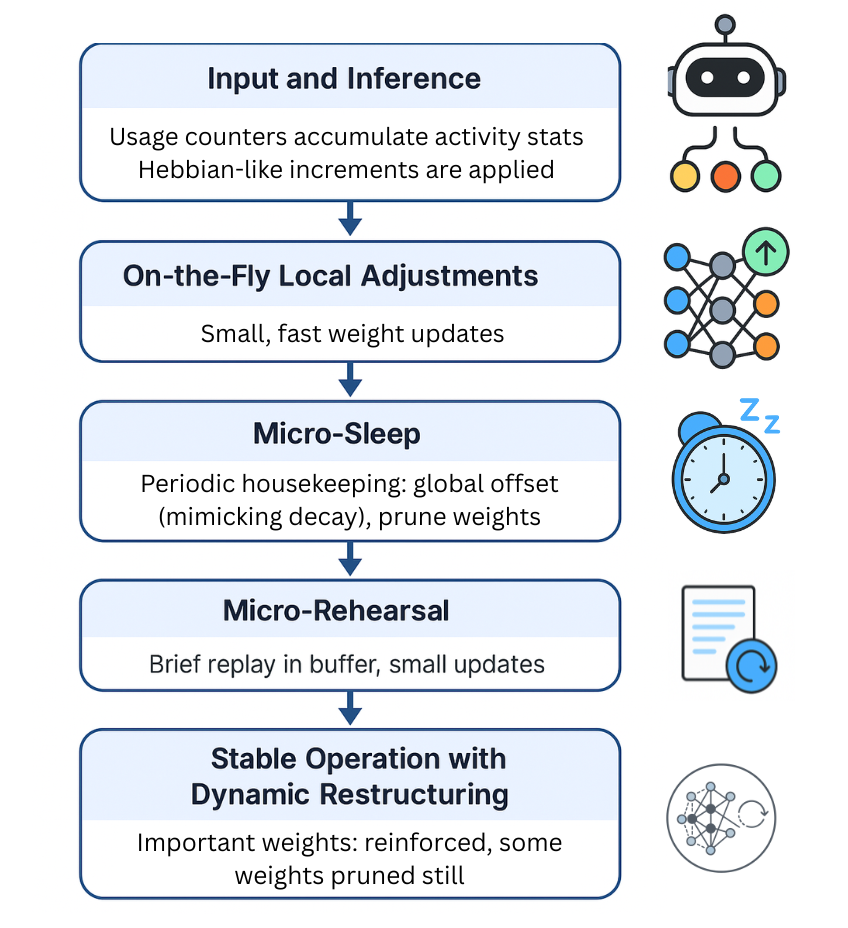}
		\caption{Timeline of operations in the Tri-Memory learning cycle, showing inference, microsleeps, rehearsal, and consolidation phases.}
		\label{fig:timeline}
	\end{figure}
	
	A typical operational cycle in our Tri-Memory system unfolds as shown in Figure~\ref{fig:timeline}. The key phases are:
	
	\begin{enumerate}
		\item \textbf{Input and Inference:}  
		The system receives new sensor input or user queries and runs a forward pass through the network. Simultaneously, usage counters accumulate activity statistics. Local Hebbian-like increments are applied to the relevant connections.
		
		\item \textbf{On-the-Fly Local Adjustments:}  
		Small, fast adjustments occur instantly after each inference. These updates reflect both correlation-driven Hebbian strengthening and minimal error-based corrections. At this stage, no large-scale decay or pruning is performed, limiting computational overhead.
		
		\item \textbf{Micro-Sleep (Periodic Housekeeping):}  
		At predefined intervals, either after a fixed number of inferences or a specified duration, the system enters a micro-sleep phase. During this brief pause, it applies a global offset to simulate synaptic decay, and selectively deactivates weights falling below the pruning threshold.
		
		\item \textbf{Micro-Rehearsal:}  
		The system momentarily revisits a few examples in its replay buffer, both recent and representative older examples, to reinforce important skills. A short backpropagation-based update corrects any drift caused by new data, ensuring stable performance over extended periods.
		
		\item \textbf{Stable Operation with Dynamic Restructuring:}  
		Over time, important weights are reinforced and remain in the model (often migrating from STM to LTM or PM), while unimportant pathways are pruned. This cyclic process ensures that the network stays dynamically optimized for the tasks at hand.
	\end{enumerate}

	\subsection{Implementation Details and Considerations}
	
	Because all weights remain non-negative in this architecture (thanks in part to ReLU-like activation functions) the pruning threshold is straightforward to manage, and negative weight handling is unnecessary.
	
	Further efficiency arises from hardware-level support for \textbf{sparse matrix multiplication}: once weights are pruned to zero, the effective size of the network shrinks, reducing forward-pass latency and energy consumption. This makes the architecture highly suitable for deployment on edge platforms with hardware acceleration for sparse computation, such as NVIDIA Jetson or similar embedded AI chips.

	\section{Conceptual Evaluation and Design Feasibility}
	
	The proposed architecture for Personalized AGI is designed to operate under real-world constraints such as limited compute, memory, and energy, particularly on edge devices. While no empirical results are presented, this section offers a comprehensive conceptual evaluation. We examine the system’s design tradeoffs, feasibility on edge platforms, and potential benefits compared to existing continual learning methods. We also discuss realistic application scenarios to illustrate how the system would operate in practice, and we conclude with open research questions for future exploration.
	
	\subsection{Theoretical Comparison with Existing Architectures}
	
	Table~\ref{tab:comparison-table} summarizes a conceptual comparison between our architecture and representative methods across continual learning, edge deployment, and brain-inspired AI. The evaluation is based on properties such as memory efficiency, catastrophic forgetting mitigation, suitability for on-device learning, and modularity for future scalability.
	
	\renewcommand\theadalign{bc}
	\renewcommand\theadfont{\bfseries}
	
	\begin{table}[htbp]
		\small 
		\centering
		\caption{Conceptual comparison of continual learning methods}
		\label{tab:comparison-table}
		\renewcommand{\arraystretch}{1.1}
		\begin{tabularx}{\textwidth}{|>{\raggedright\arraybackslash}X
				|>{\raggedright\arraybackslash}X
				|>{\raggedright\arraybackslash}X
				|>{\raggedright\arraybackslash}X
				|>{\raggedright\arraybackslash}X
				|>{\raggedright\arraybackslash}X|}
			\hline
			\textbf{Method} & 
			\textbf{Forgetting Mitigation} & 
			\textbf{Edge Compatibility} & 
			\textbf{Memory Efficiency} & 
			\textbf{Modular/ Composable} & 
			\textbf{Personal-ized Learning} \\
			\hline
			EWC / SI & Regular-ization-based & Moderate & High & Low & Weak \\
			\hline
			Replay-Based Methods & High (via rehearsal) & Low (storage-heavy) & Low & Moderate & Moderate \\
			\hline
			Progressive Nets / DEN & High (by expansion) & Low (grows over time) & Low & Moderate & Moderate \\
			\hline
			Mixture-of-Experts (MoE) & Task Isolation & Moderate & High (if sparse) & High & Weak \\
			\hline
			\textbf{Proposed Architecture} & \textbf{High (via tri-memory, pruning, replay)} & \textbf{High} & \textbf{High} & \textbf{High} & \textbf{Strong} \\
			\hline
		\end{tabularx}
	\end{table}

	Our approach uniquely combines dynamic memory allocation, lightweight Hebbian updates, nightly consolidation, and modular design, all of which mirror efficient learning strategies observed in biological brains. Unlike many prior systems that mitigate forgetting at the cost of scalability or edge deployment, our framework maintains bounded model size through pruning and selective memory retention, making it inherently suited for constrained environments.

	\subsection{Architectural Tradeoffs: Efficiency, Plasticity, and Scalability}
	
	The Tri-Memory framework addresses the stability–plasticity dilemma by design. STM handles rapid on-the-fly learning, LTM retains useful patterns through rehearsal, and PM safeguards mission-critical knowledge. The microsleep-based global decay mechanism allows continuous learning without aggressive compute use, while the separation of learning timescales prevents interference.
	
	However, the model still entails tradeoffs:
	
	\begin{itemize}
		\item \textbf{Plasticity vs. Efficiency:} High plasticity in STM is balanced by constrained update frequency and selective consolidation to avoid overfitting or over-adaptation.
		
		\item \textbf{Memory Footprint vs. Retention:} Pruning inevitably removes some older knowledge. The system counters this by emphasizing graceful forgetting, retaining high-utility memories while shedding redundant ones.
		
		\item \textbf{Scalability:} As domain complexity grows, more expert subnetworks may be needed. Although sparsely activated, these must still be managed to prevent system bloat. Future implementations may include expert merging or compression techniques.
	\end{itemize}
	
	Importantly, this architecture is hardware-aware. It assumes integration with sparse matrix acceleration and low-power compute primitives available in platforms like NVIDIA Jetson, EdgeTPUs, or even emerging neuromorphic chips. The system is built for adaptability without expanding beyond the constraints of edge-device infrastructure.
	
	\subsection{Application Scenarios as Design Validation}
	
	Three previously discussed scenarios illustrate the conceptual feasibility of this architecture:
	
	\begin{itemize}
		\item \textbf{Personal Humanoid Assistant:} Learns unique household layouts, routines, and preferences over time. Tri-memory allows initial habits to remain adaptable and later consolidated into stable behaviors. Continuous pruning and nightly replay preserve task performance without growing the model size.
		
		\item \textbf{Healthcare IoT Wearable:} Learns the user’s personal health patterns and evolves predictive insights while operating under tight energy constraints. Microsleep-based decay and sparse representation ensure only relevant signals are retained long term.
		
		\item \textbf{Smartphone Personal Assistant:} Adapts to communication style, app usage, and scheduling behaviors. Modular expert gating allows domain-specific personalization without interference across tasks.
	\end{itemize}
	
	In each case, continual learning is essential, not a bonus. The system is viable because it performs selective memory consolidation, contextual expert activation, and energy-aware housekeeping, all modeled after neurobiological learning systems.
	
	\subsection{Open Questions and Future Research Directions}
	
	While the framework offers a promising foundation for Personalized AGI on the edge, several open challenges and research opportunities remain:
	
	\begin{itemize}
		\item \textbf{Optimal Pruning Schedules and Thresholds:} How frequently and aggressively should pruning be applied to maximize both adaptability and retention? What strategies can dynamically determine the lower threshold for pruning based on usage statistics, resource constraints, or task complexity?
		
		\item \textbf{Consolidation Policies and Promotion Thresholds:} What criteria should govern promotion from STM to LTM and LTM to PM? Can upper thresholds be informed by novelty detection, performance stabilization, or user engagement signals?  
		
		\item \textbf{Catastrophic Forgetting Boundaries:} How can graceful forgetting be quantitatively measured? What minimum memory retention guarantees can be offered to prevent regressions in core functionality?
		
		\item \textbf{Expert Management:} Can redundant experts be dynamically merged or compressed to prevent modular explosion over long-term learning?
		
		\item \textbf{Neuromorphic Integration:} How might this framework be adapted to spiking neural networks or implemented on neuromorphic hardware for further energy savings?
	\end{itemize}
	
	Exploring these directions will allow this conceptual framework to evolve into a fully implemented system capable of robust, personalized AGI functionality across a wide range of devices and domains.

	\section{Discussion}
	
	Developing a personalized AGI for edge deployment raises several challenges and open questions. In this section, we discuss how our approach addresses some of these challenges and where difficulties remain. Key issues include catastrophic forgetting, memory and computation efficiency, scalability to AGI-level knowledge, and the broader implications of running a continually learning system on the edge.
	
	\subsection{Mitigating Catastrophic Forgetting}
	
	Catastrophic forgetting is the central issue in continual learning. Our architecture tackles forgetting through multiple layers: using a Tri-Memory System (so new information first goes into STM-Learner instead of directly overwriting LTM knowledge, and ultimately preserving in Permanent Memory (PM) for critical knowledge), employing replay of past data during STM training, and applying synaptic consolidation during LTM updates. Together, these create a robust ``memory stability'' net.
	
	In practice, there may still be scenarios of forgetting. For instance, if the replay memory misses some important example from a very old task, that task’s performance degrades. The graceful forgetting principles acknowledge that some forgetting might be inevitable or even necessary~\cite{golkar2019pruning}. Our approach is to prioritize important and frequently used knowledge (making it resistant to forgetting via strong consolidation), while allowing rarely used or less important details to fade. This is arguably human-like: people do forget specifics over time if they aren’t reinforced, yet core skills and frequently recalled facts stay with us.
	
	One associated challenge is concept drift. If the environment or user behavior changes gradually (over months), the AI must update its knowledge. This is not forgetting per se, but rather revising what was previously learned. Our model handles concept drift by treating it as new data that conflicts with old; the consolidation process would then adjust the LTM representations to accommodate the new concept (which might look like forgetting the old concept, but actually it’s intentional replacement). Catastrophic forgetting is only a problem if the model unintentionally loses information it still needed; distinguishing that from purposeful replacement is the key. Future research could enhance the system’s ability to autonomously judge what knowledge to keep vs. overwrite, possibly by monitoring usage frequency of certain memory items or receiving user feedback.
	
	\subsection{Memory Efficiency and Model Size Management}
	
	A continual learning system intended to operate on edge devices must carefully manage its memory and computational resources. Our architecture addresses this challenge by enforcing sparsity, applying regular pruning, and leveraging the structured design of the Tri-Memory System. The use of sparse coding ensures that only a subset of neurons are active for any given task, reducing interference and computational load. Over time, synaptic pruning reclaims capacity from rarely used connections, enabling the model to remain compact even as it accumulates new knowledge. This dynamic makes it possible for the system to scale over months or years without uncontrolled growth.
	
	Rather than attempting to store everything, the model uses a bounded, selective memory process, much like the human brain. Information that is used frequently becomes more strongly encoded and is promoted into Long-Term Memory (LTM) or the Permanent Memory (PM), while less relevant details are gracefully forgotten. This model of forgetting is not a failure but a feature, one that promotes continual adaptation while keeping the model efficient and focused.
	
	Importantly, we envision a world where humanoid agents have access to online specialized models, analogous to how humans read books or take courses when they need to learn something new. If a robot or AI assistant encounters a task beyond its current expertise, like learning a new language or understanding how to operate a B212 helicopter, it connects to a domain-specific model available online to acquire and internalize that knowledge. This mirrors how humans seek out specialized resources to learn. The model thus remains efficient on-device while still being capable of substantial knowledge expansion when needed, creating a fluid and human-like pathway to lifelong learning and specialization.
	
	\subsection{Scalability toward AGI-level Knowledge}
	
	A true AGI needs to know and do an extremely broad range of things, far beyond the basic benchmarks. As the knowledge base grows, ensuring it remains coherent is hard. If the AGI learns many disconnected tasks, will the LTM-Net end up fragmented into many experts that don’t share useful information? We hope that transfer learning will occur and the system will find commonalities and compress knowledge. We included consolidation to merge new info with old, but consolidation could be more intelligent by doing global reorganizations of knowledge (analogous to how humans sometimes have an insight that connects two domains). 
	
	Current continual learning lacks this kind of creative reorganization. It mostly preserves or adds but doesn’t globally refactor knowledge. Advancing that might be necessary for an AGI to remain efficient and general. This leans into research on representation learning and knowledge graphs within neural nets.
	
	\subsection{Privacy Considerations}
	
	Deploying a continually learning AGI on a personal device solves privacy considerations. Since learning happens on-device, user data (their behavior, conversations, etc.) need not be sent to the cloud for training, preserving privacy by design. Personalized models are kept locally. This aligns with emerging privacy-preserving AI trends.

	\subsection{Use Case Discussion}
	
	Let’s reflect on the earlier application scenarios with the perspective of our architecture and discuss any specific challenges:

	\subsubsection{Personal Humanoid Assistant}
	Imagine a humanoid robot in a home or office that assists with daily tasks. Initially, it comes with basic skills (navigation, object recognition, speech understanding trained from a generic AGI model). As it spends time with its user, it continually learns the user's routines, preferences, and environment. It learns the floor plan of the house by exploration (mapping each room, locating furniture), learns the faces and names of family members and regular visitors, and even picks up the user's particular way of giving instructions.
	
	If the user says, “Can you get me my medication?”, the robot through experience knows where the medication is usually kept and what time it’s needed, because it has formed a memory of those details. If the user teaches a new task (like watering a specific plant when its soil is dry), the robot’s STM-Learner immediately encodes the new procedure. Over time, with practice, this gets consolidated into its long-term skill set.
	
	Thanks to synaptic pruning and modular learning, the robot doesn’t become bogged down as it learns dozens of tasks. It streamlines its neural pathways, removing irrelevant ones. The continuous learning enables true personalization: this robot adapts to this household in a unique way, making it far more useful than a one-size-fits-all programmed robot. It's personality can be shaped by the household, learning the level of formality or humor the users prefer in interactions.
	
	\subsubsection{Mobile Personal Assistant (Smartphone AI)}
	A smartphone-based AI assistant could handle tasks like predictive text, scheduling, information retrieval, and entertainment recommendations in a highly personalized way. For instance, the assistant learns the user’s writing style and slang to better predict text and even auto-complete sentences as the user would phrase them. It also learns the user’s interests by observing which news they read or which music they skip, refining its recommendations daily.
	
	If the user starts learning a new language, the assistant notices increased usage of foreign phrases and adjusts its behavior (perhaps offering translations or learning to understand those phrases in context). Importantly, all this adaptation occurs on the phone, without sending the detailed usage data to cloud servers, aligning with user privacy expectations.
	
	The continual learning architecture must fit in the phone’s limited memory: this is where our model compression and expert gating help. The assistant can have separate small expert models for different tasks (text prediction, scheduling, etc.), activating and training each as needed without interfering with others. This modular approach also improves reliability: a failure or quirk learned in one domain won’t corrupt the whole system.
	
	These scenarios demonstrate a couple of applications where on-device continual learning can make AI far more effective and tailored. They also stress different aspects: some demand ultra-low power operation, others real-time learning, others long-term consistency. A generalized architecture like ours aims to be adaptable to these needs, though specific tuning would be needed per scenario.
	
	\subsection{Future Improvements}
	
	While our proposed solution makes strides, there's room for improvement and further research:
	
	\textbf{Neuromorphic Integration:} To truly capitalize on the neuroscience inspiration, future work could implement this architecture on neuromorphic hardware or event-driven frameworks. This could involve using spiking neural networks for parts of the model (particularly for Hebbian updates and sparse coding, spiking is a natural fit).
		
	\textbf{Theoretical Guarantees:} Continuous learning algorithms currently lack strong theoretical guarantees. Developing theory for stability–plasticity trade-off, or error bounds for a model that is pruned and updated in an online fashion, would provide deeper understanding. Our architecture could be studied theoretically by modeling it as an online learning problem with memory constraints.
		
	\textbf{Generalization vs. Personalization Balance:} An interesting line of inquiry is how a personal edge AGI can maintain general reasoning abilities (so it doesn’t become too narrow). If it only ever sees one user’s data, it might overfit to that user’s world. Humans mitigate this by being exposed to diverse situations especially early on (education). Perhaps a personal AI would periodically need a dose of “general experience” (maybe via simulation or a base model update from a central server covering broad data) to stay broadly capable. Achieving this balance is key: being specialized but also adaptable to entirely new domains as needed.
	
	While many pieces of the puzzle are in place (thanks to extensive research in continual learning and neuroscience-driven AI), integrating them into a coherent, scalable, and efficient AGI system is an ongoing endeavor. Our proposed architecture is a step in that direction, but extensive experimentation and iteration is needed to refine these ideas into a robust real-world system.
	
	While empirical evaluation remains a future step, this framework provides a biologically grounded and system-level perspective to inform the development of on-device AGI.

	\section{Conclusion}
	
	This paper explores the vision of Personalized AGI via Neuroscience-Inspired Continuous Learning Systems, charting a path toward AI systems that learn and adapt throughout their lifetime while operating on edge devices. We began with a survey of the literature in continual learning, highlighting how the AI community has tackled catastrophic forgetting and the emerging intersection with neuroscience principles. We examined how concepts like synaptic consolidation (from neuroscience) map onto algorithms like EWC and SI, how memory replay in AI is analogous to the brain’s experience replay, and how structural plasticity (growing or pruning neurons) provides a route to lifelong model adaptation. This review underscores that no single technique is sufficient; instead, a combination of strategies, like the brain’s multifaceted approach, is needed.
	
	Delving into neuroscience, we discussed four key principles: (1) Synaptic pruning, which guides us to continually optimize and compress models to retain efficiency; (2) Hebbian plasticity, which offers mechanisms for rapid learning of associations; (3) Sparse coding, which encourages using distributed but sparse representations to minimize interference and resource use; and (4) Dual memory systems, which directly inspired our architectural separation of fast and slow learning components.
	
	Building on those insights, we proposed a novel AI architecture for continual learning suitable for edge deployment. The architecture integrates a fast learning module (for quick, plastic updates) with a stable knowledge base (that accumulates and preserves knowledge), coupled with consolidation process, replay, pruning, and expert gating. It leverages advanced model compression so that the system remains deployable on devices with limited hardware. The design is intended to be modular and extensible, allowing future improvements such as more sophisticated expert networks or neuromorphic hardware implementations to plug in. While the architecture is ambitious, it is grounded in existing research findings and technologies. Each component has at least a prototype in prior work (e.g., dual-memory in meta-learning, pruning in continual learning, MoE for efficient inference, etc.), which gives confidence that the integration is feasible.
	
	We outlined a conceptual evaluation and design feasibility, emphasizing the importance of both learning performance and efficiency on real or simulated edge setups. Continual learning benchmarks, combined with on-device tests, can demonstrate whether the system achieves the desired balance: high plasticity with low forgetting, all within constrained resources.
	
	In discussing the challenges and future directions, we acknowledged that catastrophic forgetting, while mitigated, can never be entirely eliminated. Rather, it must be managed. We highlighted that the stability–plasticity trade-off will remain a central consideration, and methods that dynamically adjust this trade-off (potentially with inspiration from how different brain regions or neuromodulators control learning rates) are a promising area. We also pointed out the significance of bounded resources, which means an AGI must know how to forget or compress old information in an intelligent way to make room for new, a process that our pruning and module replacement aims to handle. Scalability to true AGI remains a grand challenge: our approach can be seen as a stepping stone, but reaching human-level breadth and depth of knowledge will likely require further breakthroughs in representation learning and possibly new paradigms of computation.
	
	From an application standpoint, the move toward on-device learning AI is already visible in industry trends (e.g., personalized keyboards, adaptive smart cameras). Our work pushes this toward a more general and autonomous learning capability. The scenario examples illustrate that if realized, personalized continual learning AGIs could provide more intuitive, responsive, and privacy-conscious technology, whether it’s a robot that becomes part of the family or a phone that genuinely knows its user. These are compelling motivators for continuing research in this direction.
	
	Future research will likely explore hybrid approaches that combine our on-device learning with occasional cloud collaboration (to share knowledge among agents or get heavy-duty computation when needed). Another exciting avenue is more deeply neuroscience-integrated models: for example, using brain-inspired memory circuits (like differentiable neuro-modulated working memory) or even direct neurophysiological analogs (like models that mimic sleep phases for consolidation). On the hardware side, co-designing algorithms with emerging memory technologies (as Kudithipudi et al. advocate) could unlock orders-of-magnitude efficiency improvements, making lifelong learning AI not just a software achievement but a hardware revolution too.
	
	In conclusion, achieving personalized AGI on the edge is a complex, multi-disciplinary challenge. By learning from the brain’s lifelong learning strategies and by advancing continual learning algorithms, we can make significant strides. This paper has presented an integrative approach that, while requiring extensive validation, sets the stage for a new generation of AGI, ones that learn every day, adapt to us, and do so on devices we own. Such agents would mark a shift from static AI models to ever-evolving companions, aligning AI behavior more closely with the fluid and cumulative nature of human learning. We hope this work provides a useful roadmap for researchers and practitioners aiming to bring us closer to that vision.

\newpage
\bibliographystyle{elsarticle-num}
\bibliography{references}

\newpage
\section*{Author biographies}

\newcommand{\authorbox}[3]{%
	\noindent
	\begin{minipage}[t]{0.22\textwidth}
		\includegraphics[width=\linewidth]{#1}
	\end{minipage}%
	\hfill
	\begin{minipage}[t]{0.74\textwidth}
		\raisebox{75pt}[0pt][0pt]{\parbox[t]{\linewidth}{\textbf{#2} #3}}
	\end{minipage}
	\vspace{3em}
}

\authorbox{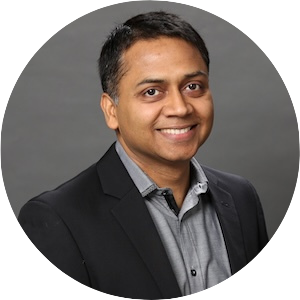}{Rajeev Gupta} {
	is an entrepreneur with extensive experience in AI strategy, risk modeling, and intelligent systems. He is the Co-Founder of Cowbell, where he leads innovation at the intersection of AI and Risk. Rajeev is also passionate about robotics, with a strong interest in applying AI to enable real-world autonomy.
}

\authorbox{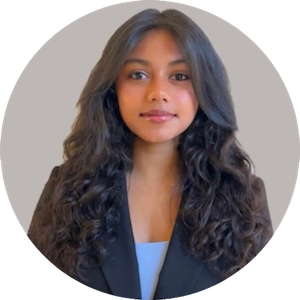}{Suhani Gupta}{
	is the Founder and CEO of the Neuro Health Alliance, where she leads work at the intersection of mental health advocacy and neuroscience-inspired AI. Her focus on cognitive modeling and personalized systems informed the integration of biological principles into the architecture proposed in this paper.
}

\authorbox{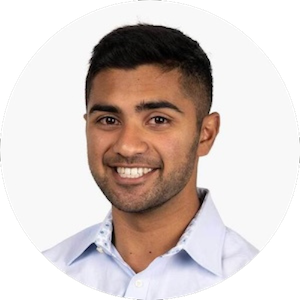}{Ronak Parikh}{
	is an AI Engineer based in New York with experience in deep learning, MLOps, and deploying production-ready models. His work centers on generative architectures and structured approaches for building adaptive, real-world AI systems, contributing to the conceptual design of this paper's framework.
}

\authorbox{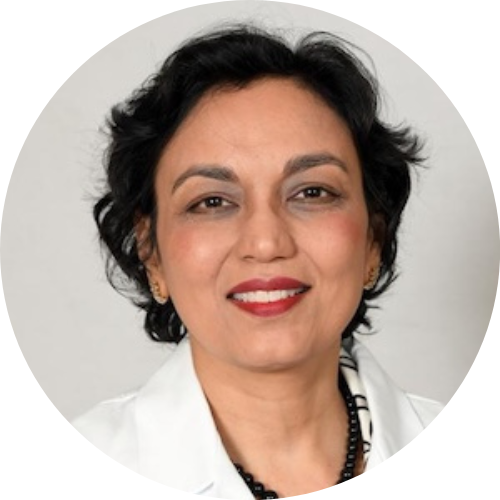}{Dr. Divya Gupta}{
	is the Medical Director of the JFK Neuroscience Sleep Center at HMH. A neurologist with deep expertise in neuroscience and sleep medicine, her understanding of neuroplasticity, memory consolidation, and synaptic pruning directly informed the biological foundations of this paper’s continual learning framework.
}

\authorbox{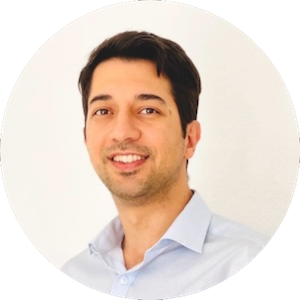}{Amir Javaheri}{
	is the Head of Data Science and AI at Cowbell, where he leads initiatives in LLMs and applied AI. His work focuses on building scalable AI systems and translating research into robust, production-ready solutions. His expertise shaped the practical implementation considerations in this paper's architecture.
}

\authorbox{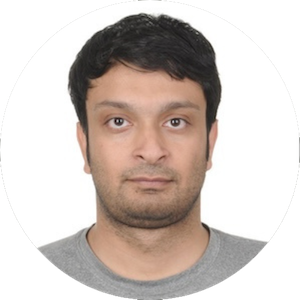}{Jairaj Singh Shaktawat}{
	is a Staff AI Engineer with expertise in transformers, deep learning, and model optimization. His work on efficient inference and representation learning informed the architecture’s memory compression and edge deployment strategies discussed in this paper.
}

\end{document}